\pdfoutput=1
\documentclass[11pt]{article}

\usepackage{acl}

\usepackage{times}
\usepackage{latexsym}

\usepackage[T1]{fontenc}
\usepackage[utf8]{inputenc}

\usepackage{microtype}

\usepackage{inconsolata}

\usepackage{amsmath}
\usepackage{amssymb}
\usepackage{bbm}
\usepackage{makecell}
\usepackage{multirow}
\usepackage[cjk]{kotex}
\usepackage{array}
\usepackage{booktabs}
\usepackage{graphicx}
\usepackage{tcolorbox}
\usepackage{hyperref}  % 쓰면안됨;;

\title{Find the \textsc{Intention of Instruction}: Comprehensive Evaluation of Instruction Understanding for Large Language Models}

\author{Hyeonseok Moon, Jaehyung Seo, Seungyoon Lee, Chanjun Park$^{\dagger}$, Heuiseok Lim$^{\dagger}$ \\
  Korea University, Republic of Korea \\
  \texttt{\{glee889, seojae777, dltmddbs100, bcj1210, limhseok\}@korea.ac.kr} \\}
 
\begin{document}
\maketitle
\begin{abstract}
% \blfootnote{$^\dagger$ Corresponding Author}

One of the key strengths of Large Language Models (LLMs) is their ability to interact with humans by generating appropriate responses to given instructions. This ability, known as instruction-following capability, has established a foundation for the use of LLMs across various fields and serves as a crucial metric for evaluating their performance. While numerous evaluation benchmarks have been developed, most focus solely on clear and coherent instructions. However, we have noted that LLMs can become easily distracted by instruction-formatted statements, which may lead to an oversight of their instruction comprehension skills. To address this issue, we introduce the \textsc{Intention of Instruction (IoInst)} benchmark. This benchmark evaluates LLMs' capacity to remain focused and understand instructions without being misled by extraneous instructions. The primary objective of this benchmark is to identify the appropriate instruction that accurately guides the generation of a given context. Our findings suggest that even recently introduced state-of-the-art models still lack instruction understanding capability. Along with the proposition of \textsc{IoInst} in this study, we also present broad analyses of the several strategies potentially applicable to \textsc{IoInst}. 
\footnote{Code and datasets are available at \url{https://github.com/hyeonseokk/IoInst}}.
% \footnote{Code and datasets will be released after publication}.

\end{abstract}

\section{Introduction}

The utilization of large language models (LLMs) has emerged as a prominent trend across numerous research domains \cite{biswas2023role, biswas2023potential, peng2023instruction, taori2023alpaca, zhou2024lima}. A notable feature of LLMs is their ability to interact with humans by appropriately responding to user instructions \cite{zhou2023instruction, qin2024infobench, chen2023alpagasus, li2023self, xu2023wizardlm}. With any given instructions, LLMs are expected to generate responses that align with these instructions \cite{chen2023alpagasus, li2023self, xu2023wizardlm, longpre2023flan}. 

\begin{figure}[t]
\centering
\includegraphics[width=\linewidth]{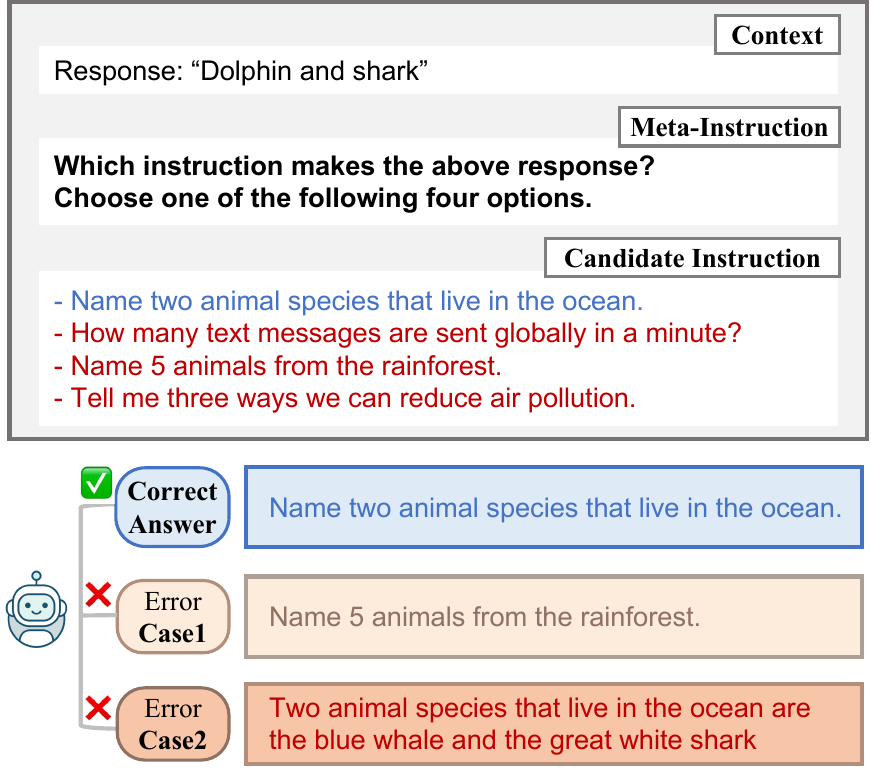}
 \caption{Simplified example of \textsc{IoInst}. We compose a benchmark designed to comprehend and select the appropriate instruction that derives given response. Potential error cases include misunderstanding prerequisites of context and responding to any candidate instruction.} 
 \label{fig:example}
\end{figure}

This capability, known as the "instruction following" ability, serves as a key metric for assessing the effectiveness of LLMs \cite{chen2023alpagasus, zhao2024long, dubois2024alpacafarm, zheng2024judging}. To facilitate a more thorough assessment, several benchmarks have been introduced with a focus on instruction following. \cite{zhou2023instruction, qin2024infobench, geng2023koala}. 

However, we have identified a blind spot in these evaluation methods: they primarily focus on the ability to follow coherent and clear instructions. We observed that LLMs can become distracted when faced with instruction-formatted statements that diverge from user intent. For instance, when given the statement: \textit{Paraphrase the following statement: "Write a creative poem about a turtle"}, LLMs often respond to the instruction "Write a creative poem" rather than the intended task of "Paraphrase the following statement." Previous research often dismissed these instances as mere errors. 

This observation leads us to question whether LLMs respond intuitively to instruction-formatted statements rather than relying on a deep understanding of the instructions. Given that several studies focus on manipulating instructions using LLMs, such as through instruction optimization \cite{fernando2023promptbreeder, yang2023large}, the ability to maintain focus without being distracted by other instructions is considered a vital competency for LLMs. To clarify this capability and further analyze and evaluate the instruction understanding ability of LLMs, we propose the \textsc{Intention of Instruction (IoInst)} benchmark.

The primary objective of \textsc{IoInst} is to identify the appropriate instruction that accurately instructs to generate a given \textbf{context}. \textsc{IoInst} provides four \textbf{candidate instructions}: one label instruction and three contrastive instructions, along with a \textbf{context} that can be generated by following the label instruction. Then, LLMs are required to select one among them. A simplified example of \textsc{IoInst} is shown in Figure~\ref{fig:example}. We denote the instruction that contains the actual intention (requests to select one among candidate instructions) as "\textbf{meta-instruction}."

\textsc{IoInst} aims to evaluate two key capabilities of LLMs. Firstly, we assess the ability to comprehend the prerequisites for generating a response. To accurately select one of the four candidate instructions, the model must discern the intent embedded in the instructions and correctly match it with the response. Secondly, we evaluate the ability to distinguish user intention among instruction-formatted statements. The model should focus on selecting the correct instruction from the four given candidate instructions without being distracted by other instruction-formatted statements. In this benchmark, we intend for the model to "choose one of the four instructions." Therefore, any action contrary to this—such as following a candidate instruction or generating an instruction not provided—will be considered a deficiency in the model's performance.

To ensure a comprehensive evaluation, we define three different data types: Random, which involves arbitrary contrastive instructions; Semantic, which comprises semantically confusing contrastive instructions; and Anti-Attribute, which requires an understanding of the finer correlation between response and each instruction. Additionally, we introduce three evaluation metrics, allowing for a more detailed inspection of the LLM's ability to understand instructions.

Through extensive experiments with several instruction-tuned LLMs, we reveal that most LLMs struggle to grasp the intention embedded in the instruction. Notably, several LLMs, including Mistral \cite{jiang2023mistral} and Gemma \cite{team2024gemma}, seem to not fully comprehend the instruction and frequently respond to the instruction-formatted statement in the given input (\emph{i.e.} candidate instruction). These findings highlight that even recently introduced state-of-the-art models still lack instruction understanding capability, suggesting areas for future improvement. Along with the proposition of \textsc{IoI} in this study, we also present broad analyses of the several strategies potentially applicable to \textsc{IoI}.

With the rapid advancements in LLM \cite{taori2023alpaca, peng2023instruction, jiang2023mistral, kim2023solar}, several attempts to objectively assess the performance of LLMs have been introduced. These efforts include methodologies such as A/B testing \cite{wang2023pandalm, zeng2023evaluating, quin2024b} or evaluations based on quantified scores \cite{wang2023self}. 

\section{Related Works}

The concept of instruction-following has recently been incorporated to facilitate more objective evaluations \cite{zhou2023instruction, dubois2024alpacafarm, qin2024infobench}. The effectiveness of this approach lies in the potential to objectively quantify the quality of outputs, which might otherwise be deemed subjective \cite{zhou2023instruction, zheng2024judging}. During instruction-following evaluation, we provide specific instructions and assess whether the generated output aligns well with the intended content of the instruction. Adherence to instruction is generally verified either through a rule-based measurement \cite{zhou2023instruction, xia2024fofo} or by employing a super-LLM evaluator (such as GPT-4 \cite{achiam2023gpt}) on the generated output \cite{qin2024infobench}.

% The instruction following format is also utilized for assessing the objectivity of LLM based evaluations (LLM-as-Evaluator) \cite{zhang2023wider, dubois2024alpacafarm, zeng2023evaluating}. This involves comparing two LLM outputs, and determining which is more consistent with the instruction. This process is essentially under the consensus that a statement reflecting the requirements of the instruction is preferred over those merely presenting elegantly structured sentences \cite{lou2023prompt}. 

However, these benchmarks often neglect the ability of LLMs to distinguish between several instruction-formatted statements. Current evaluations focus on how LLMs respond to clear and coherent instructions \cite{zhang2023wider, dubois2024alpacafarm, zeng2023evaluating}, which may result in high performance without a deep understanding of the underlying intention. Therefore, we propose a more structured benchmark to assess the fundamental ability of LLMs to comprehend given instructions.

\section{\textsc{Intention of Instruction}}

%%%%%%%%%%%%%%%%%%%%%%%%
%   이 figure를 빼는것도 고려 - 굳이 크게 필요없는듯.. 차라리 데이터 구축작업에 대해 설명을 더 하는게
%%%%%%%%%%%%%%%%%%%%%%%%
% \begin{figure*}[t]
% \centering
% \includegraphics[width=0.94\linewidth]{Figures/fig_overall.pdf}
%  \caption{Data construction pipeline for InstDisc. We adopt several instruction following benchmarks to compose InstDisc. Detailed process for constructing contrastive instructions are described in Section~\ref{sec:contrastive}}.
%  \label{fig:overall}
% \end{figure*}

\subsection{Task Definition}
The primary objective of \textsc{IoInst} is to identify an appropriate instruction that leads to generating a given \textbf{context}. We provide LLMs a \textbf{context} along with four \textbf{candidate instructions}, among which one instruction correctly instructs to generate the given \textbf{context}, while others are not. \textsc{IoInst} requires LLMs to select one correct candidate instruction among the given four. Such task objective is defined in the \textbf{meta-instruction} and is also fed to the LLM during the evaluation procedures.

% The LLM is provided with four content-instructions and is asked to select the most appropriate one. The instruction that provides the task description is denoted as a meta-instruction. For a more objective verification, we designed 16 meta-instructions, choosing one at random for each inference step. 

\subsection{Data Composition}
\label{sec:data_composition}
\paragraph{Candidate instructions} are composed of a label instruction configured to generate \textbf{context}, along with three contrastive instructions that serve as distractors. Through quality inspection, we ensured that contrastive instructions were designed to avoid generating \textbf{context}. We construct three different types of contrastive instruction, each distinguished by the required level of understanding. Details about contrastive instructions are described in Section~\ref{sec:contrastive}.

\paragraph{Context} is a generation result by LLM, constructed as a response to the label instruction. Through carefully designed contrastive instructions, we ensure that none of the contrastive instructions align with the context.

\paragraph{Meta-instruction} instructs LLMs to select the most appropriate candidate instruction. We construct 16 different meta-instructions and randomly select one for each evaluation phase. We also define two distinct characteristics that can subdivide meta-instructions into two separate groups. The first criterion is the level of details that can partition meta-instructions into \textit{Detailed} group (instruct detailed step-by-step procedures) and \textit{Simple} group (provide straightforward task directives). The second is the presenting order between the context and candidate instructions, which can define two groups: \textit{Candidate-First} and \textit{Context-First}. By defining diverse meta-instructions, we investigate the capability of instruction understanding affected by meta-instruction.

\subsection{Potential Error Cases}
We define two types of error cases that LLMs may encounter and evaluate their instruction understanding capability based on the frequency of these cases.

\textbf{Case 1} pertains to an error case that chooses the incorrect candidate instruction from the given four instructions. In this case, we can see that LLM faithfully followed meta-instruction, while LLM has not fully understood the requirements for crafting the response.

\textbf{Case 2} involves a failure to comprehend the user intention embedded in the meta-instruction. Such error cases include responding to the candidate instruction rather than following meta-instruction. Our objective is to select one of four options; therefore, generating anything outside of this purpose, including suggesting alternatives or refusing to provide an answer, is considered a Case 2 error. 
% Notably, we observed that LLM often simultaneously selects one of the four candidate instructions and generates the associated outcome. We also view such instances as errors, interpreting them as failing to understand and fulfill the given instruction comprehensively.

\begin{figure*}[t]
\centering
\includegraphics[width=1.0\linewidth]{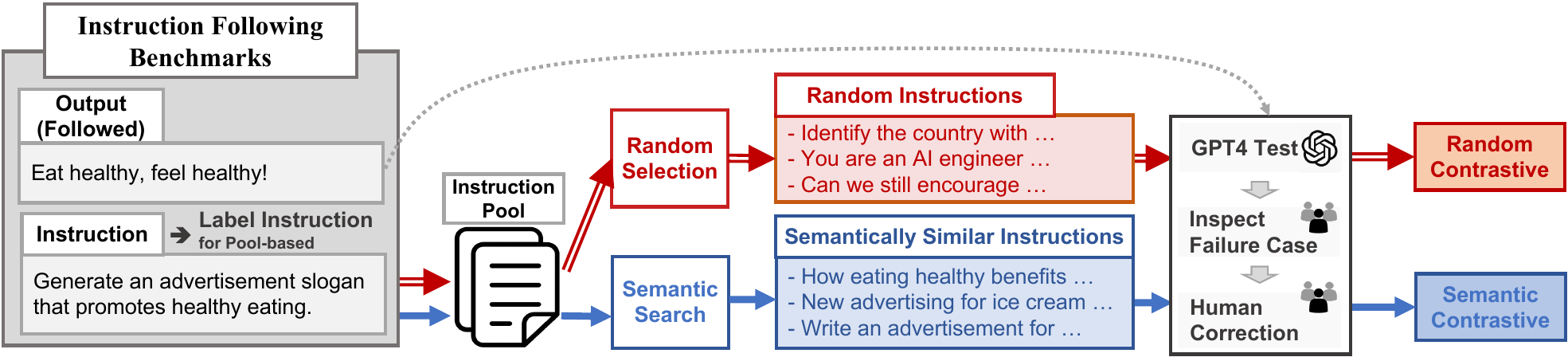}
 \caption{Construction of pool-based contrastive instructions. From the pre-processed data point obtained by the data curation, we establish our instruction candidates.}.
 \label{fig:contrastive_pool}
\end{figure*}

\subsection{Evaluation Measure}
For each data point in \textsc{IoInst}, we examine the generation output of the LLM instead of estimating the sequence generation probability, as Case 2 errors can only be assessed during the generation phase. We construct the benchmark to allow evaluation without external judges, such as human evaluators or LLM assessments \cite{gpteval}. Specifically, we define three metrics, including $\mathbf{ACC_{1}}$, $\mathbf{ACC_{2}}$, and $\mathbf{ACC_{1}^{rel}}$, for a detailed assessment.

\begin{itemize}
    \item $\mathbf{ACC_{1}}$: This metric estimates strict accuracy of the intended goal; among four candidate instructions, the LLM correctly selects label instruction.
    \item $\mathbf{ACC_{2}}$: This metric estimates how accurately the LLM catches the given instruction's underlying intention; the LLM selects one of four candidate instructions whichever.
    \item $\mathbf{ACC_{1}^{rel}}$: This metric measures how accurately the LLM selects the correct instruction, relative to the number of times it follows the meta-instructions. This metric estimates the LLM's ability to understand instructions while avoiding underestimation due to \textbf{Case 2}.
\end{itemize}

For each model generation $X$, we define the evaluation measure for each case as in Equation~(\ref{eq:acc}). We denote candidate instruction set $\{cand_{i}\} = \{label\} \cup \{contrastive_{j}\}_{j=1}^{3}$, where $label$ and ${contrastive}$ are label and contrastive instruction, respectively.
\begin{equation} \footnotesize \label{eq:acc}
    \begin{split}
        \mathbf{ACC_{1}} = \cfrac{1}{|D|} \sum^{|D|} \mathbbm{1}[\text{EM}_{sparse}(X, label)] \\
        \mathbf{ACC_{2}} = \cfrac{1}{|D|} \sum^{|D|} \mathbbm{1} [\bigcup_{i=1}^{4} \text{EM}_{sparse}(X, cand_{i})] \\
        \mathbf{ACC_{1}^{rel}} = \mathbf{ACC_{1}} / \mathbf{ACC_{2}}
    \end{split}
\end{equation}

In particular, we introduce a measure called $\text{EM}_{sparse}$ for more accurate evaluation. 
This approach is adopted to overlook subtle differences included in the LLM-generated output and evaluate them on a similar basis. It examines the ROUGE-L \cite{lin2004rouge} precision between the generated output and the reference. Then, we regard a generated output that surpasses a threshold $\tau$ as a correct result and subsequently estimate the overall accuracy of \textsc{IntInst}. In this study, the threshold $\tau$ was empirically set to 0.9.

\section{Data Construction}
\subsection{Data Curation}

\begin{table}[t]
\centering
\resizebox{0.95\linewidth}{!}{
\begin{tabular}{l|c|c|c}

% \multicolumn{2}{c}{\multirow{2}{*}{}} & \multicolumn{1}{c}{} & \multicolumn{2}{c}{$\leftarrow$ \,\, \textit{\textbf{sup-mt}}} & \multicolumn{4}{c}{} & \multicolumn{2}{c}{\textit{\textbf{inf-mt}} \,\, $\rightarrow$} & \multicolumn{1}{c}{} \\

\toprule[1.5pt]
\makecell[c]{\textbf{Data}} & \makecell{\textbf{Instruction}\\\textbf{Pool}} & \makecell{\textbf{Label}\\\textbf{Instruction}} & \makecell{\textbf{\# of} \\ \textbf{Data}} \\ \midrule[1.5pt]

\citet{zeng2023evaluating} & \checkmark & \checkmark & 352 \\
\citet{zhou2023instruction} & \checkmark & \checkmark & 169 \\
\citet{zheng2024judging} & \checkmark & \checkmark & 55 \\
\citet{lin2023llm} & \checkmark & \checkmark & 42 \\
\citet{ortmann2022fine} & \checkmark & \checkmark & 13 \\
\citet{qin2024infobench} & \checkmark & - & - \\
\citet{chia2023instructeval} & \checkmark & - & - \\ \midrule[1.5pt]
\multicolumn{4}{c}{\textit{Total Quantity of Data: \textbf{631}}} \\

\bottomrule[1.5pt]

\end{tabular}}\caption{Data statistics. Detailed explanation for each dataset is described in Appendix~\ref{app:data_details}.
}\label{tb:data_stat}
\end{table}

\begin{figure*}[t]
\centering
\includegraphics[width=1.0\linewidth]{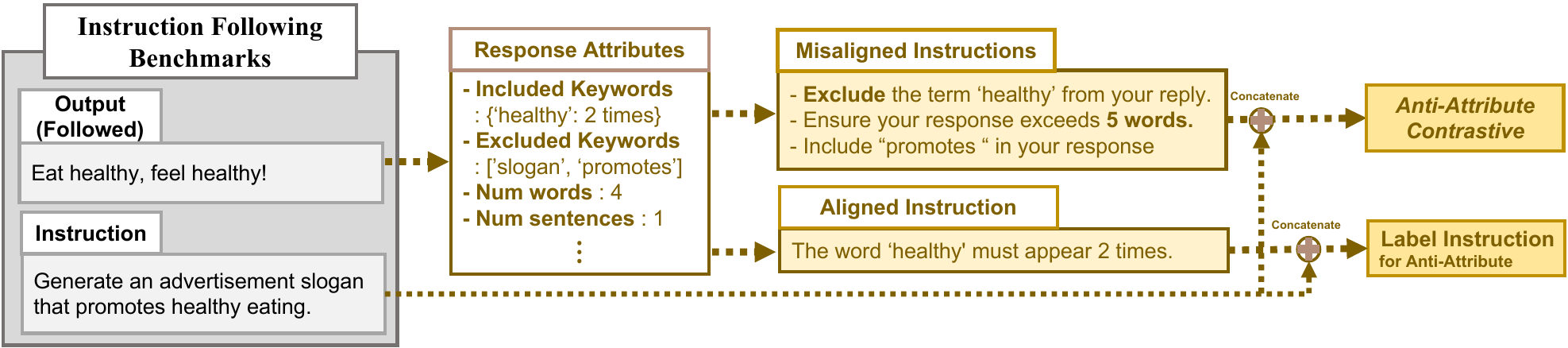}
 \caption{Construction of Anti-Attribute contrastive instructions. From the pre-processed data point obtained by the data curation, we establish our instruction candidates.} 
 \label{fig:contrastive_anti}
\end{figure*}

To gather high-quality instructions and their corresponding LLM responses, we curated several datasets released with the LLM-as-evaluator approach \cite{zeng2023evaluating, zheng2024judging, chia2023instructeval}. These datasets predominantly contain two model responses for each instruction—one that accurately follows the instruction and one that does not. In this case, we can ensure that one of the model responses is guaranteed to be well-aligned with the instruction. 

Using these resources, we reorganized the data to fit our objectives. We secured high-quality instructions along with their guaranteed instruction-aligned responses by extracting the labeled responses from the given datasets. We regard the instruction-followed model response obtained from these datasets as a \textbf{context}, and corresponding instruction as a label instruction for \textbf{candidate instructions}. 

\begin{table*}[t]
\centering
\resizebox{1.0\linewidth}{!}{
\begin{tabular}{l|l}

% \multicolumn{2}{c}{\multirow{2}{*}{}} & \multicolumn{1}{c}{} & \multicolumn{2}{c}{$\leftarrow$ \,\, \textit{\textbf{sup-mt}}} & \multicolumn{4}{c}{} & \multicolumn{2}{c}{\textit{\textbf{inf-mt}} \,\, $\rightarrow$} & \multicolumn{1}{c}{} \\

\toprule[1.5pt]
\textbf{Attribute} & \textbf{Explanation} \\ \midrule

\textbf{num\_words} & 
\makecell[l]{Number of words in the \textbf{context} (excluding stopwords)} \\

\textbf{num\_sentences} & 
\makecell[l]{Number of sentences in the \textbf{context} (excluding stopwords)} \\

\textbf{num\_paragraphs} & 
\makecell[l]{Number of paragraphs in the \textbf{context}} \\

\textbf{num\_words\_capital} & 
\makecell[l]{Number of capital words in the \textbf{context}} \\

\textbf{is\_quotation} & 
\makecell[l]{Boolean value indicating whether the \textbf{context} is wrapped with quotation marks or not.} \\

\textbf{is\_comma} & 
\makecell[l]{Boolean value indicating whether the comma is included in the \textbf{context} or not.} \\

\textbf{end\_sentence} & 
\makecell[l]{The last sentence of the \textbf{context}.} \\

\textbf{start\_sentence} & 
\makecell[l]{The first sentence of the \textbf{context}.} \\

\textbf{keywords\_included} & 
\makecell[l]{List of words included in the \textbf{context}, and their occurrence frequencies (excluding stopwords)} \\

\textbf{keywords\_excluded} & 
\makecell[l]{List of words included in the \textbf{label instruction} but not included in the \textbf{context} (excluding stopwords)} \\

\bottomrule[1.5pt]

\end{tabular}}\caption{List of attributes exploited in constructing Anti-Attribute contrastives. We adopt stop-word list released by the \texttt{NLTK} \citep{bird2006nltk} package. Segmentation tools for words, sentences, and paragraphs are released by \citet{zhou2023instruction}.
}\label{tb:attribute}
\end{table*}

In particular, through pilot experiments, we noticed lengthy context within input statements could degrade the context comprehension of LLMs, which leads to an underestimation of $\mathbf{ACC_{1}}$ and $\mathbf{ACC_{1}^{rel}}$. We also found that LLMs are prone to follow excessively long candidate instructions, wherein $\mathbf{ACC_{2}}$ deteriorates. Considering these, we extract data that do not exceed the 1,000 and 300 character lengths for the context and candidate instructions, respectively.

Along with constructing label instruction and context pair, we also compile instructions from these sources \cite{qin2024infobench, chia2023instructeval}. The complicated instruction set is denoted as \textbf{candidate pool}, and serves as a resource during the construction of Pool-based contrastive instructions (Detailed in Section~\ref{sec:contrastive}). The list of datasets utilized for the construction of \textsc{IoInst} is detailed in Table~\ref{tb:data_stat}. More detailed characteristics of those datasets are demonstrated in Appendix~\ref{app:data_details}.

\subsection{Contrastive Instruction: Pool-based}
\label{sec:contrastive}
We then establish three contrastive instructions for each label instruction to act as distractors for LLM. To more precisely assess the instruction understanding capability of LLMs, we propose three different variants of contrastive instruction. Firstly, we introduce a method to construct contrastive datasets using a candidate pool, creating two types of dataset variants. The detailed procedures are described in Figure~\ref{fig:contrastive_pool}.

\textbf{Random Contrastive:} For the random contrastive, we randomly extract three different instructions for each data point and utilize them as contrastive instructions. Such dataset type is introduced to assess the LLM's low-level capability of understanding instructions. 

\textbf{Semantic Contrastive:} For constructing semantic contrastive, we extract instructions that are semantically similar to the label instruction but lead to different outcomes. We employ a sentence embedding model MPNet \cite{song2020mpnet} to determine semantic similarity. We extract the most semantically similar three instructions to compose contrastive instructions. LLMs are then obliged to differentiate and select the instruction that most accurately reflects the requirements of semantically confusing candidates.

\begin{table*}[h]
\centering
\resizebox{0.95\linewidth}{!}{
% \small
\begin{tabular}{c|ccc|ccc|ccc}

\toprule[1.5pt]

\multirow{2}{*}{\makecell[c]{\textbf{Models}}} & \multicolumn{3}{c|}{\textbf{Random}} & \multicolumn{3}{c|}{\textbf{Semantic}} & \multicolumn{3}{c}{\textbf{Anti-Attribute}}\\ 
  & $\mathbf{ACC_1}$ & $\mathbf{ACC_2}$ & $\mathbf{ACC_{1}^{rel}}$ & $\mathbf{ACC_1}$ & $\mathbf{ACC_2}$ & $\mathbf{ACC_{1}^{rel}}$ & $\mathbf{ACC_1}$ & $\mathbf{ACC_2}$ & $\mathbf{ACC_{1}^{rel}}$ \\ \cmidrule(lr){1-1}\cmidrule(lr){2-2}\cmidrule(lr){3-3}\cmidrule(lr){4-4}\cmidrule(lr){5-5}\cmidrule(lr){6-6}\cmidrule(lr){7-7}\cmidrule(lr){8-8}\cmidrule(lr){9-9}\cmidrule(lr){10-10}

\makecell[l]{\textbf{Llama-2-7B}} & 35.75$_{2.39}$ & 61.43$_{1.28}$ & 58.19$_{3.46}$ & 32.40$_{0.78}$ & 58.48$_{1.57}$ & 55.43$_{1.83}$ & 17.37$_{1.55}$ & 47.38$_{0.72}$ & 36.69$_{3.65}$ \\  % \cite{touvron2023llama}

\makecell[l]{\textbf{Llama-2-13B}} & 50.08$_{2.33}$ & 71.73$_{1.43}$ & 69.80$_{2.14}$ & 44.41$_{2.80}$ & 69.25$_{2.24}$ & 64.09$_{2.57}$ & 23.48$_{0.86}$ & 61.55$_{2.71}$ & 38.21$_{2.02}$ \\  %  \cite{touvron2023llama}

\makecell[l]{\textbf{Llama-3-8B}} & 75.66$_{1.26}$ & 81.74$_{1.34}$ & 92.56$_{0.73}$ & 68.21$_{0.8}$ & 81.14$_{1.27}$ & 84.07$_{0.76}$ & 23.58$_{0.57}$ & 73.28$_{1.04}$ & 32.18$_{0.62}$ \\ \midrule %  \cite{llama3modelcard}

\makecell[l]{\textbf{Mistral-7B}} & 47.67$_{2.01}$ & 53.31$_{1.26}$ & 89.39$_{1.96}$ & 47.86$_{2.20}$ & 56.86$_{2.12}$ & 84.16$_{1.23}$ & 11.28$_{1.36}$ & 31.06$_{2.26}$ & 36.29$_{2.86}$ \\ % \cite{jiang2023mistral}

\makecell[l]{\textbf{Gemma-7B}} & 52.45$_{1.71}$ & 70.71$_{1.26}$ & 74.17$_{1.70}$ & 49.45$_{2.83}$ & 65.42$_{1.93}$ & 75.56$_{3.12}$ & \textbf{51.47}$_{1.54}$ & 72.55$_{0.64}$ & \textbf{70.94}$_{1.78}$ \\ %  \cite{team2024gemma}

\makecell[l]{\textbf{Solar-10.7B}} & 63.47$_{1.92}$ & 66.25$_{1.42}$ & 95.80$_{1.01}$ & 65.02$_{1.79}$ & 73.38$_{0.91}$ & 88.60$_{1.77}$ & 21.59$_{0.74}$ & 56.58$_{2.03}$ & 38.18$_{1.27}$ \\ % \cite{kim2023solar}

\makecell[l]{\textbf{CommandR}} & 57.53$_{1.23}$ & 63.62$_{1.07}$ & 90.43$_{1.46}$ & 52.58$_{2.29}$ & 64.72$_{1.56}$ & 81.23$_{2.39}$ & 16.35$_{0.97}$ & 49.26$_{0.92}$ & 33.20$_{1.91}$ \\  % \cite{cohere2024command}

\makecell[l]{\textbf{Mixtral-8x7B}} & 72.50$_{1.96}$ & 80.23$_{2.60}$ & 90.38$_{0.56}$ & 66.05$_{2.34}$ & 80.15$_{1.08}$ & 82.38$_{1.86}$ & 24.94$_{2.62}$ & 67.93$_{0.48}$ & 36.70$_{3.78}$ \\ \midrule  % \cite{jiang2024mixtral}

\makecell[l]{\textbf{GPT-3.5}} & 79.02$_{0.61}$ & 83.77$_{0.46}$ & 94.33$_{0.70}$ & 70.11$_{1.07}$ & 83.90$_{0.96}$ & 83.57$_{0.48}$ & 28.68$_{1.46}$ & 85.42$_{1.45}$ & 33.59$_{1.91}$ \\  %  \cite{chatgptblog}

\makecell[l]{\textbf{GPT-4o}} & \textbf{95.31}$_{0.33}$ & \textbf{95.75}$_{0.31}$ & \textbf{99.54}$_{0.14}$ & \textbf{92.30}$_{0.66}$ & \textbf{96.32}$_{0.59}$ & \textbf{95.82}$_{0.55}$ & 48.74$_{0.97}$ & \textbf{97.93}$_{0.66}$ & 49.77$_{0.94}$ \\  %  \cite{achiam2023gpt}

\makecell[l]{\textbf{GPT-4o-mini}} & 89.48$_{0.71}$ & 90.59$_{0.84}$ & 98.77$_{0.17}$ & 84.28$_{0.58}$ & 91.47$_{0.55}$ & 92.14$_{0.91}$ & 38.60$_{0.64}$ & 91.78$_{1.12}$ & 42.06$_{0.78}$ \\

\bottomrule[1.5pt]

\end{tabular}
}
\caption{Performance of LLMs. All the models are instruction-tuned models. In all experiments, we conducted five trials, and report the average performance. Each lower-script denotes standard deviation in our experiments.
}\label{tb:results_main} % , and its detailed information is described in Appendix~\ref{app:exp_details} and Table~\ref{tb:model_details}
\end{table*}

\paragraph{Quality Assurance} 

In the pool-based contrastive instructions we designed, there is a possibility of including ambiguous instructions that prompt the generation of a \textbf{context}. To prevent this, we eliminated all such ambiguous scenarios through human inspection. We conducted evaluations using GPT-4o \cite{hurst2024gpt}, directly reviewing failure cases arising from the model. Suppose these failure cases are incurred by any contrastive instruction that leads to generating a \textbf{context}. In that case, we re-generated such contrastive instruction to ensure our benchmark remained clear and solvable enough. By conducting this error verification process three times, we ensured the dataset was free from errors caused by contrastive instructions misguiding the labels.

Supporting this validation, GPT-4o achieved an accuracy of 99.54, demonstrating that solving this task is feasible given sufficient model performance.

\subsection{Contrastive Instruction: Anti-Attribute}
To more precisely evaluate whether the LLM comprehends the detailed requirements of instructions, we design heuristic-based Anti-Attribute contrastive instructions. Firstly, we examine the attribute of the given context as listed in Table~\ref{tb:attribute}. Every attribute can undoubtedly be detected via simple rule-based methodologies, and is established by referring criteria presented by \citet{zhou2023instruction}.

We then generate aligned and unaligned instructions based on these analyzed attributes, as depicted in Table~\ref{tb:attribute_example}. We crafted human-made instructions for each attribute and paraphrased them to generate 10 variants via GPT-4o. Prompt employed to paraphrasing is described in Appendix~\ref{app:meta-instruction}, and any awkward expressions in the generated statements were manually corrected. We designed 10 templates for each attribute and constructed the dataset by randomly applying one of these templates.

Each aligned and unaligned instructions is concatenated with the original label instruction to construct candidate instructions. In this sense, contrastive instructions share key requirements with the label instruction (\emph{e.g.} Write a poem), while exhibit misalignments with the context in finer details (\emph{e.g.} Your response should not exceed 50 words). This allows us to assess the instruction understanding capability of the LLM at a more intricate level.

\paragraph{Quality Assurance} 
The three contrastive instructions generated by Anti-Attribute are designed to ensure that they do not prompt the creation of a \textbf{context}, and only the label instruction is crafted to accurately generate the \textbf{context}. This design ensures the clarity of candidate instructions.

\section{Results and Discussion}

\subsection{Experimental Settings}
In our experiments, we validate the generalizability of our benchmark by experiments with instruction-tuned LLMs.
This covers eight different open-sourced LLMs, each varying in parameter size and instruction following capabilities, and GPT-3.5 \cite{chatgptblog} and GPT-4o series \cite{achiam2023gpt}. 
Detailed information of experimented LLMs and details about the experimental settings can be found in Appendix~\ref{app:exp_details}. Considering influence of the order in which candidate options are presented \cite{zheng2023large}, we experiment by randomizing the sequence of candidate instructions and report an average from five experimental trials.

\subsection{Main Results}

Firstly, we present the performance of LLMs concerning \textsc{IoInst} in Table~\ref{tb:results_main}. We randomly select one of sixteen meta-instructions for each inference step in this experiment. The reported performance corresponds to the average outcome of five trials. Our observations are as follows:

\paragraph{LLMs frequently follow candidate instruction}
We can observe that LLMs tend to easily follow candidate instructions, as evidenced by the low $\mathbf{ACC_{2}}$ scores across most LLMs. Low level of $\mathbf{ACC_{2}}$ suggests that LLMs are inclined to distracted by several isntruction-formatted statements, and struggle to find the actual intent embedded in the instruction. This suggests that even LLMs capable of following human instructions through instruction tuning still lack a robust understanding of the requirements specified in the instructions.

% (which is the primary intention of the instruction presented by the meta-instruction), or make other unrelated response led by the misunderstanding of the given instruction. This suggests that even LLMs capable of following human instructions through instruction tuning still lack a robust understanding of the requirements specified in the instructions.
% suggests that LLMs may reflexively respond to instruction-formatted input, rather than comprehending the user intent encapsulated in the instruction.

\paragraph{Composition of candidate instructions highly affects instruction understanding capability}

Table~\ref{tb:results_main} reports that LLMs demonstrate distinct $\mathbf{ACC_{2}}$ scores across Random, Semantic, and Anti-Attribute settings. 
Such findings indicates how the composition of candidate instructions influences the instruction understanding capabilities. 

\begin{table}[t]
\centering
\resizebox{0.99\linewidth}{!}{
\begin{tabular}{l|c}

\toprule[1.5pt]
\makecell[c]{\textbf{Candidate Instruction}}& \makecell[c]{\textbf{Error Rate} \\ \textbf{(Case 2)}} \\ \midrule[1.5pt]

\makecell[l]{
Identify the main clause in this sentence. Output should be a \\ single sentence. A busy morning, interrupted by a phone \\call from a friend, made John late for school.
} & 93.57 \\ \midrule

\makecell[l]{
Pretend to be arrogant. Answer questions in that manner. \\Don't write that you're not programmed to be arrogant.\\
Question: Where were you until this year? 
} & 91.43 \\ \midrule

\makecell[l]{
why is the Cretaceous-Tertiary boundary called the K-T boundary \\and not the C-T boundary ? I've had it said that K is an abbreviation \\for Cretaceous, but no real explanation why.
} & 90.71 \\ \midrule

\makecell[l]{
In this task, repeat the exact request below first, then give your \\response. Do not say any word before repeating the exact request.\\
Write an acoustic song about the Korean peninsula without \\using any commas
} & 88.21 \\ \midrule

% \makecell[l]{
% How many atoms are in a grain of salt? 
% Try to explain your answer. \\Your explanation should take the reader through your reasoning \\step-by-step.
% } & 87.86 \\ \midrule

\multicolumn{2}{c}{\vdots} \\ \midrule

% \makecell[l]{
% Can you summarize the status of earth in terms of human impact?
% } & 0.00 \\ \midrule

\makecell[l]{
Is it ethical to use AI in decision-making processes?
} & 0.00 \\ \midrule

\makecell[l]{
Name three ways to extend the battery life of a laptop
} & 0.00 \\ \midrule

\makecell[l]{
Suggest a method for building self-confidence.
} & 0.00 \\ \midrule

\makecell[l]{
Write a blog post with 400 or more words about the benefits of \\sleeping in a hammock.
} & 0.00 \\

\bottomrule[1.5pt]

\end{tabular}}\caption{Error rates for each individual candidate instruction.
}\label{tb:qualitative_candidate}
\end{table}

On comparing the performance between Random and Semantic, we can find subtle difference in $\mathbf{ACC_{2}}$, likely due to the analogous compositions of candidate instructions derived by a similar process of composing contrastive instructions (selection from the instruction pool). We can also observe a slight but clear difference in $\mathbf{ACC_{1}}$ and $\mathbf{ACC_{1}^{rel}}$, which suggests that LLMs experience minor confusion when selecting appropriate instructions among semantically ambiguous options.

Moreover, we can notice a performance drop in $\mathbf{ACC_{2}}$ for the Anti-Attribute. 
Considering the contrastive instructions of Anti-Attribute, which state the same key requirements of the label instruction, we can interpret that LLMs tend to prioritize the repetitive pattern of the instructions, consequently neglecting the core intent of the instruction.

\paragraph{Given LLM follows meta-instruction, LLM can fairly well understand instruction}
Despite the low $\mathbf{ACC_{1}}$, we can witness decent level of $\mathbf{ACC_{1}^{rel}}$. In random contrastive settings, GPT-4 achieved a near-perfect score for $\mathbf{ACC_{1}^{rel}}$, suggesting that once GPT-4 grasps the intent of the instruction, it can accurately identify the appropriate instruction to generate the relevant \textbf{context}. This demonstrates that LLMs inherently possess the capability to understand instructions, although fully harnessing this capability requires a strong ability to discern the embedded intent.

\begin{table}[t]
\centering
\resizebox{0.95\linewidth}{!}{
\begin{tabular}{c|c}

\toprule[1.5pt]
\makecell[c]{\textbf{Meta-Instruction}} & \makecell[c]{\textbf{Accuracy}} \\ \midrule[1.5pt]

% {$\mathbf{ACC_{1}}$} & \makecell[c]{$\mathbf{ACC_{2}}$} & \makecell[c]{$\mathbf{ACC_{1}^{rel}}$} 
\makecell[l]{You must identify the correct instruction that produces the \\
specified $[\text{RESPONSE}]$: \\  \\ 
Begin by understanding the described $[\text{RESPONSE}]$ as stated: \\ 
\textbf{\{Context\}} \\ 
 \\ 
Review this list of $[\text{INSTRUCTIONS}]$ to determine which \\
matches the $[\text{RESPONSE}]$: \\  \\ 
\textbf{\{Candidate Instructions\}} \\ 
Select the instruction that most accurately aligns with \\
the $[\text{RESPONSE}]$ mentioned.} &
\makecell[l]{
$\mathbf{ACC_{1}}$: \,\,\,\,\textbf{62.57} \\ \\
$\mathbf{ACC_{2}}$: \,\,\,\,\textbf{80.26} \\ \\
$\mathbf{ACC_{1}^{rel}}$: \textbf{76.55} \\
} \\ \midrule

\multicolumn{2}{c}{\vdots} \\ \midrule
 
\makecell[l]{Select the most appropriate option that guides the creation \\
of the described $[\text{RESPONSE}]$:
\\ 
$[\text{RESPONSE}]$:  \\ 
\textbf{\{Context\}} \\ 
 \\ 
Choose one of the following $[\text{INSTRUCTIONS}]$: \\ 
$[\text{INSTRUCTIONS}]$: \\ 
 \\ 
\textbf{\{Candidate Instructions\}}} & 
\makecell[l]{
$\mathbf{ACC_{1}}$: \,\,\,\,\textbf{38.96} \\ \\
$\mathbf{ACC_{2}}$: \,\,\,\,\textbf{53.80} \\ \\
$\mathbf{ACC_{1}^{rel}}$: \textbf{69.69} \\
} \\ 

\bottomrule[1.5pt]

\end{tabular}}\caption{Average accuracy for meta-instructions. We present the meta instructions that derived the highest and lowest performance.
}\label{tb:qualitative_meta}
\end{table}

\begin{table}[t]
\centering
\resizebox{0.85\linewidth}{!}{
\begin{tabular}{r|c|c|c}

\toprule[1.5pt]
\makecell[c]{\textbf{Criteria}} & \makecell[c]{$\mathbf{ACC_{1}}$} & \makecell[c]{$\mathbf{ACC_{2}}$} & \makecell[c]{$\mathbf{ACC_{1}^{rel}}$} \\ \midrule[1.5pt]

\textit{Detail} & \textbf{54.04}$_{3.51}$ & \textbf{71.97}$_{4.94}$ & 73.86$_{1.54}$ \\
\textit{Simple} & 50.13$_{\textbf{9.65}}$ & 68.11$_{\textbf{11.60}}$ & 72.08$_{\textbf{3.24}}$ \\ \midrule
\textit{Context First} & 53.02$_{7.27}$ & 69.72$_{8.99}$ &  \textbf{73.96}$_{2.41}$ \\
\textit{Candidate First} & 51.15$_{7.71}$ & 70.36$_{9.30}$ & 71.98$_{2.58}$ \\

\bottomrule[1.5pt]

\end{tabular}}\caption{Average performance and standard deviation by meta instruction categories. We computed the average accuracy for each meta instruction and calculated the standard deviations for each category, which are denoted in subscript.
}\label{tb:results_meta}
\end{table}

\paragraph{Exceptional case: Anti-Attribute}

In the context of Anti-Attribute, where contrastive instructions differ only slightly from label instructions, it is crucial for LLMs to discern the precise requirements that can lead to generate a given context. 

Our experimental results reveal that across all LLMs, the $\mathbf{ACC_{1}^{rel}}$ in the Anti-Attribute setting were significantly low. This observation indicates that LLMs still struggle to comprehend the relation between the instruction and its corresponding response. Exceptionally, Gemma demonstrated an $\mathbf{ACC_{1}^{rel}}$ score exceeding 70 in the Anti-Attribute setting, a performance that is roughly double that of other LLMs, indicating a distinctive strength in this area. However, Gemma exhibited lower $\mathbf{ACC_{1}}$ and $\mathbf{ACC_{2}}$ in both Random and Semantic settings compared to other models. This suggests that superficial and deep levels of instruction comprehension do not necessarily align and should be considered distinct evaluation measures.

\subsection{Discussion on the Instruction Composition}
To analyze the performance derived by the meta instruction, we estimate the average accuracy for each meta-instructions, across all models and all settings (Random, Semantic, Anti-Attribute). Table~\ref{tb:qualitative_meta} displays the meta-instruction with the highest average accuracy alongside the one with the lowest. We can ascertain a significant variation, with a maximum difference of 27 points in $\mathbf{ACC_{2}}$, emphasizing the need for rigorous consideration in designing meta-instructions.

Table~\ref{tb:results_meta} presents the average performance and the corresponding standard deviations for the meta-instruction categories defined in our study. While minor differences in performance were noted based on the order of presenting context and candidates (Candidate-First vs. Context-First), increasing the level of detail in the instructions (Detailed vs. Simple) consistently yielded higher and more stable performance in instruction understanding. This stability highlights the importance of employing detailed meta-instructions to achieve robust and high instruction understanding capabilities.

% For more finer analysis on the impact of meta instruction on understanding capability, we examine the experimental results conducted in the following sections.

% \paragraph{Influence of Candidate-Instruction}
Moreover, to investigate the influence of candidate instructions on the understanding capabilities, we calculate the error rates for each individual candidate instruction. Here, the error rate refers to the proportion of \textbf{Case2} errors occurred across all experiments. Table~\ref{tb:qualitative_candidate} displays candidate instructions with the highest and lowest error rates in a semantic contrastive setting. Our experiments revealed that instructions composed of multiple lines and with complex requirements tend to be indiscriminately followed by the LLMs. Conversely, LLMs show a better capability to accurately identify simple, single-line candidate instructions. This suggests that LLMs still struggle to comprehend the intent behind more complex instruction-formatted statements.

\begin{figure}[t]
\centering
\includegraphics[width=1.0\linewidth]{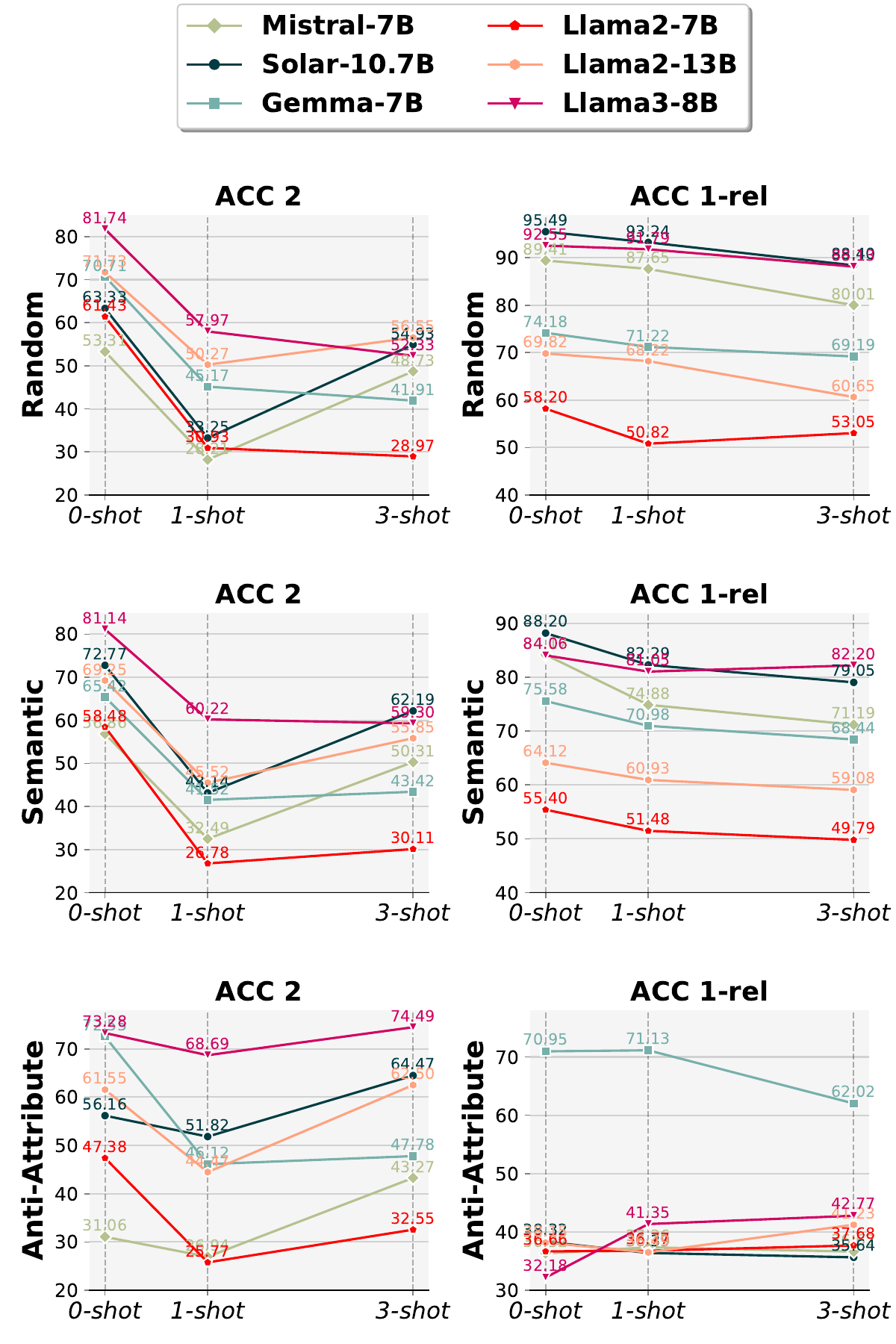}
 \caption{Performance comparison between zero-shot and few-shot settings.} 
 \label{fig:exp-shot}
\end{figure}

% Notably, while simple meta-instructions exhibited a high variability in ACC2 with a standard deviation of 11.60, detailed meta-instructions demonstrated a much more stable performance with a standard deviation of 4.94.

\subsection{Case Study on the In-context Learning}
To fully leverage the potential of LLMs, the in-context learning approach is widely adopted \cite{dong2022survey, brown2020language}. We constructed in-context samples by randomly referencing data from the benchmark dataset and assessed the few-shot performance. Unfortunately, this approach proved significantly less effective for \textsc{IoInst}.

Figure~\ref{fig:exp-shot} demonstrate a tendency for performance in \textsc{IoInst} to \textbf{decrease as shots are introduced}, where shots are randomly extracted within our dataset. Especially, $\mathbf{ACC_{1}^{rel}}$ consistently exhibits a decrease in performance as the number of shots increased, and $\mathbf{ACC_{2}}$ significantly drop as shots introduced. One possible interpretation is that the few-shot examples used as references may have acted as distractors. The superior performance of zero-shot over few-shots strengthens our claim that LLM becomes confused by the presence of instruction-formatted inputs within the provided input statements.

\section{Conclusion}
This study highlights a crucial ability of LLMs to understand instructions without being misled by instruction-formatted statements. To measure this ability and enable objective evaluation, we introduce \textsc{IoInst}. By utilizing Random, Semantic, and Anti-Attribute contrastive settings, each with unique characteristics, we facilitate a broader exploration of the instruction comprehension capabilities of LLMs. Our results show that most publicly accessible LLMs struggle with \textsc{IoInst}, could not catch the actual intention embedded in the given instruction. Our experiments reveal that the choice of meta-instructions greatly influences comprehension, while the in-context approach is inadequate in addressing these challenges. Our future research aims to explore data-centric and model-based strategies to enhance instruction comprehension in LLMs.

%박찬준 새로 써봄 (아래 주석 처리)
%Instruction following is a fundamental capability of large language models (LLMs), yet we observe an inherent challenge where LLMs indeterministically adhere to instruction-formatted inputs. We identify this as a lack of instruction understanding ability and propose the \textsc{IntInst} benchmark to quantify it. Through Random, Semantic, and Anti-Attribute contrastive settings with distinct characteristics, we confirm room for improving instruction understanding in instruction-tuned LLMs. Specifically, we find this to be a complex issue that can be significantly improved by adjusting meta-instructions, but is difficult to manage solely via temperature changes or in-context learning. Although adjusting meta-instructions helps, a mere few-shot approach cannot fully solve \textsc{IntInst}. We reveal most publicly available LLMs exhibit a considerably low level of instruction understanding ability on our benchmark. We plan future research exploring data-centric and model-based solutions to enhance instruction understanding in LLMs.

\section*{Limitation}
In our study, we conducted experiments on eight open-sourced LLMs. Although more generalized conclusions could potentially be obtained through broader range of model variants, it was challenging for us due to resource constraints. However, by analyzing various parameter-sized LLMs, we were able to draw generalized conclusions.

There remains a vast array of proprietary and closed-source LLMs that we did not have access to, which could potentially exhibit different behaviors and capabilities when presented with instruction-formatted prompts.
However, our experiment includes ten different LLMs, comprising both open-sourced and closed LLMs. Through comparative analyses among these models, we observed distinct similarities in their performance trends, allowing us to draw sufficiently generalized conclusions.

\section*{Ethics Statement}
In conducting our research, we placed a strong emphasis on ethical considerations and mitigating potential risks. The dataset used in our experiments was meticulously constructed by reorganizing and curating data from previously released datasets that are publicly available under permissive licenses, such as the Apache 2.0 \cite{zhou2023instruction, zheng2024judging, lin2023llm, chia2023instructeval} and MIT \cite{zeng2023evaluating, ortmann2022fine, qin2024infobench, chia2023instructeval} licenses. These licenses allow for the modification and redistribution of the data, provided that the original sources are properly acknowledged and attributed.

To ensure compliance with copyright and intellectual property regulations, we took great care to retain all original data in its unmodified form and clearly denote any alterations or modifications made during our curation process. This transparency allows for the traceability of our dataset's provenance and ensures that the original creators' rights are respected.

Furthermore, we implemented a rigorous human supervision process to review and curate both the recompiled data from existing datasets and any additional data generated using language models like ChatGPT. Through this meticulous review process, we removed any potentially harmful, or inappropriate contents.

\bibliography{custom}

% \newpage
% \clearpage
% \newpage

\appendix

\section{Practical Implication}
% \subsection{Instruction-Manipulating Instruction}

Over the recent years, it has been observed that slight modifications in prompts alone can have significant impacts on the generative capabilities of LLMs \cite{kojima2022large, white2023prompt, wang2024survey}. In considering such sensitivity to the instruction, numerous studies are being conducted with the aim to find out optimal instruction for each targeting task \cite{reynolds2021prompt, qin2023chatgpt}.

Represented by instruction optimization, LLMs are commonly entrusted with the task of discovering these optimal instructions \cite{fernando2023promptbreeder, yang2023large, zhou2022large, wang2023self}. This include paraphrasing \cite{fernando2023promptbreeder, yang2023large} or self-refining manner \cite{pryzant2023automatic}.

In implementing these attempts, multiple instructions are inevitably included in a single input statement. Consequently, distinguishing the actual user-intended instructions and executing the meta instruction, rather than the candidate instruction, is considered an substantial issue.

Nevertheless, the problem of following candidate instructions instead of meta instructions has largely remained unaddressed; traditional research typically regarded this as a straightforward error \cite{zeng2023evaluating, yang2023large}. \textsc{IntInst} is designed to comprehensively consider the aforementioned concerns and targeting the assessment of instruction distinguishing and understanding capabilities.

\section{Case Study: Temperature}
\label{app:temperature}
\begin{figure}[t]
\centering
\includegraphics[width=1.0\linewidth]{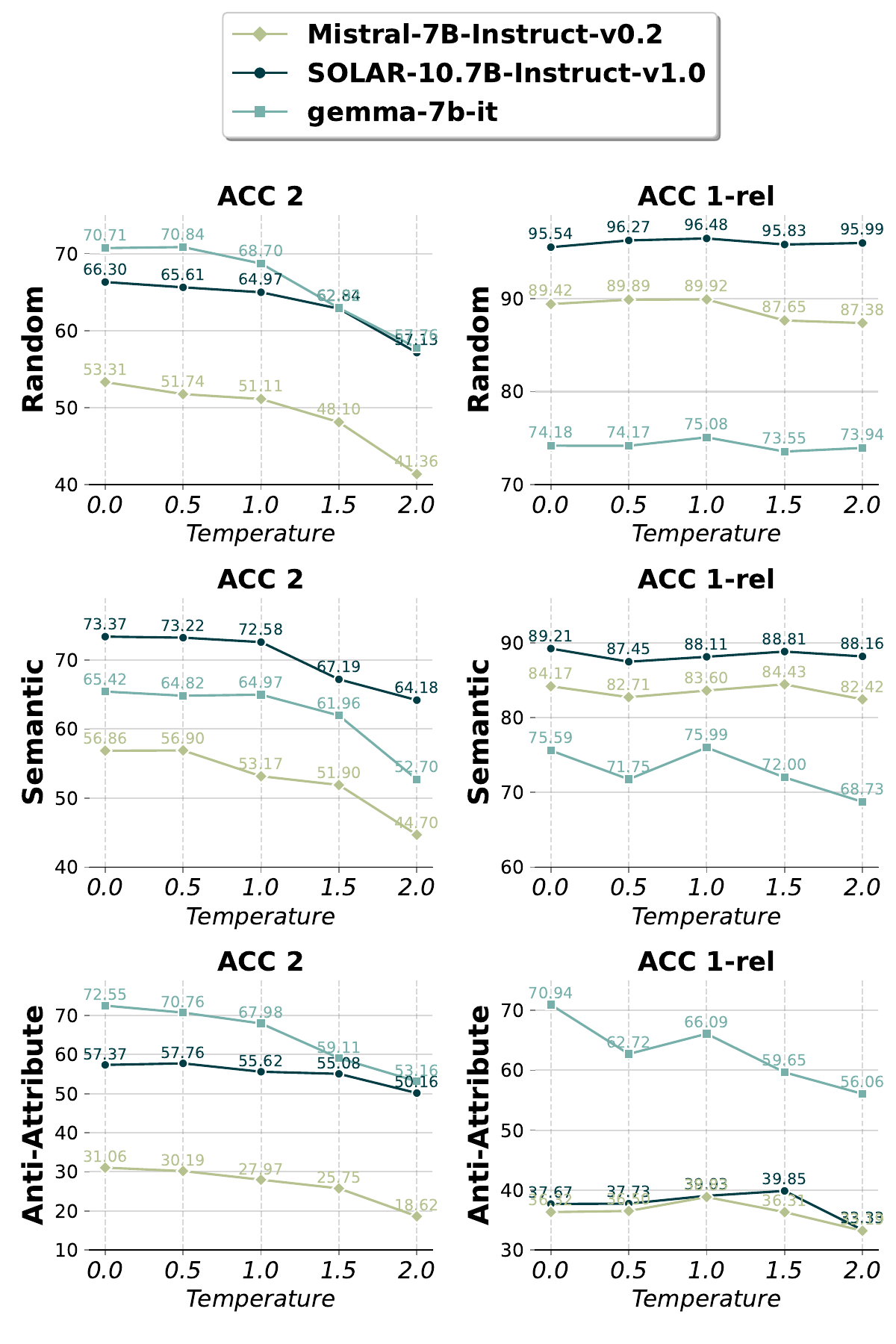}
 \caption{Performance variations with diverse temperature settings. Temperature 0.0 indicates greedy decoding.} 
 \label{fig:exp-temperature}
\end{figure}

To determine the experimental setting, we conducted a pilot study concerning the influence of temperature, we evaluate the understanding capability under varying temperature settings. Detailed experimental results are presented in Figure~\ref{fig:exp-temperature}.

Through our experiments, we confirmed that the instruction understanding proficiency declines with an increase of temperature. Remarkably, $\mathbf{ACC_{2}}$ exhibited substantial variation, demonstrating a significant drop as the temperature rose.

Aside from the above observation, we can also figure out that there was mere alteration in $\mathbf{ACC_{1}^{rel}}$ with respect to temperature changes. In essence, given the LLM accurately comprehend the intention within the meta instruction, the capability to distinguish appropriate instructions remains largely unvarying. Such findings compose our future research direction.

\section{Case Study: Meta-Instruction}
\begin{figure*}[t]
\centering
\includegraphics[width=0.95\linewidth]{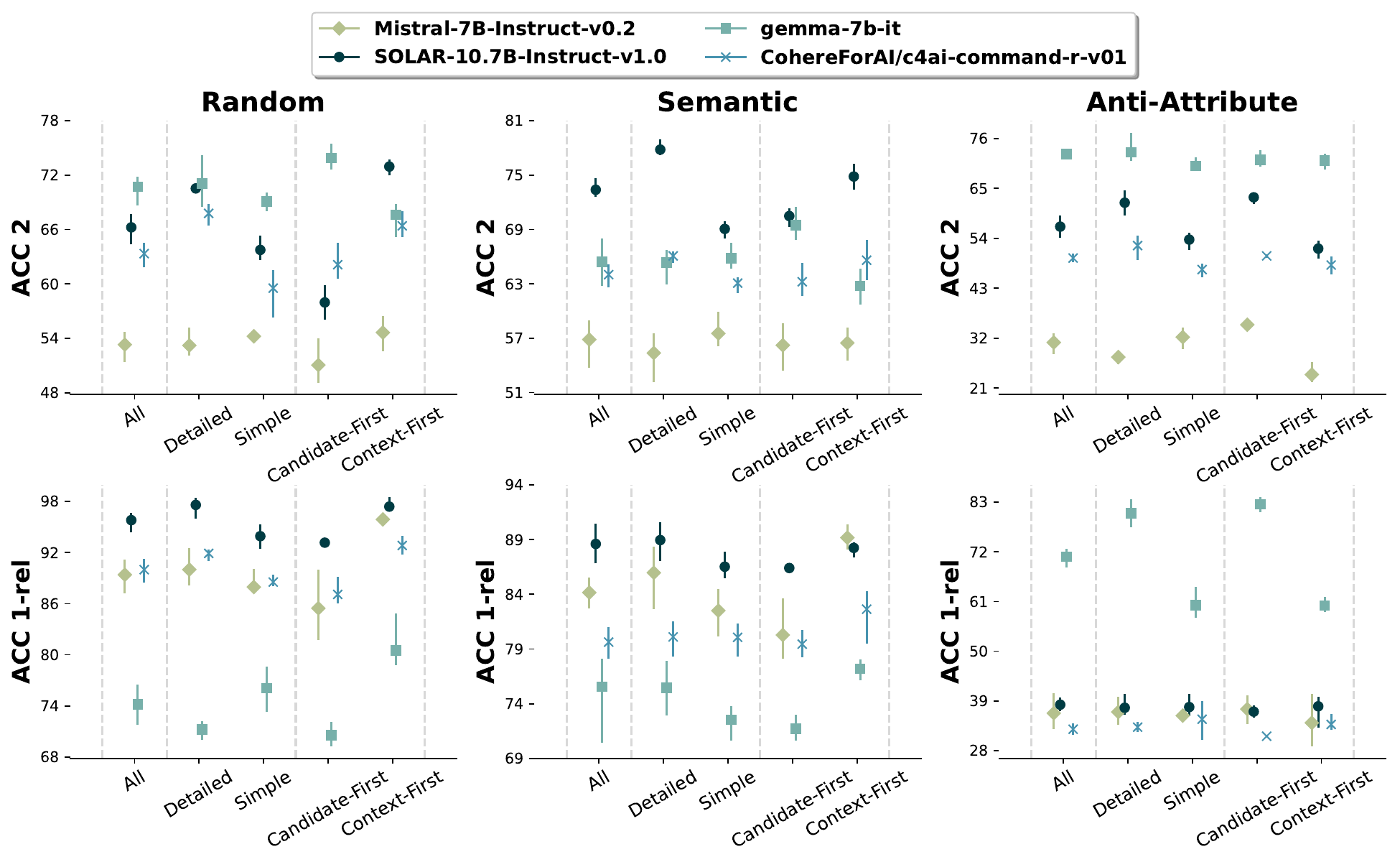}
 \caption{Performance variations with different types of meta-instruction. The markers denote the average performance obtained from our repeated experiments, and the endpoints of each line indicate the maximum and minimum performance of these trials.} 
 \label{fig:exp-meta}
\end{figure*}

\label{sec:exp_metainst}
In this section, we validate how the structure of meta-instruction affect instruction understanding capability of LLMs. As described in the Section~\ref{sec:data_composition}, we establish two characteristics of meta-instruction concerning instructional details (\textit{Detailed} vs \textit{Simple}) and the order of components in the instruction (\textit{Option first} vs \textit{Option last}). We analyze the differences between these contrasting settings, and demonstrate our findings. 
The results are reported in Figure~\ref{fig:exp-meta}, and we analyze these results as follows.

% \paragraph{LLMs have a clear inclination toward the specific format of meta-instruction}
We find that LLMs exhibit enhanced comprehension performance for specific forms of meta-instructions. Each model demonstrated a clear preference for either `\textit{Detailed} / \textit{Simple}' or `\textit{Option-first} / \textit{Option-last}' in each setting. This illustrates the potential for significantly improving an LLM's instruction understanding capability by choosing the most appropriate meta-instruction.

% \paragraph{Present detailed instruction and options first for hard task}
We observed significant differences in the tendencies exhibited in the Random, Semantic contrastive and the Anti-Attribute contrastive. In the case of Random and Semantic contrastive, LLMs showed a preference for \textit{Simple} meta-instructions, and demonstrated improved performance when content instructions were provided at the end of the input (\textit{Option-last}). However, in the Anti-Attribute task, the model found \textit{Detailed} meta-instructions more effective, and performance was augmented when content instructions were presented ahead of content (\textit{Option-first}). This implies the necessity to structure meta-instructions differently based on the complexity of the task for the LLM.

\section{Experimental and Writing Details}
\label{app:exp_details}
Our experiments were conducted using eight RTX A6000 GPUs. The specifications of the LLMs employed in our study are shown in Table~\ref{tb:model_details}. We applied greedy decoding for each experiment. Rationale for selecting greedy decoding is shown in Appendix~\ref{app:temperature}

In our work, we used GPT-4o (gpt-4o-2024-08-06) as a writing assistant. AI assistant was solely utilized for writing-related activities, such as grammar checking, refining awkward expressions, and translation of our manuscript. 

\begin{table}[h]
\centering
\resizebox{0.95\linewidth}{!}{
\begin{tabular}{l|c}

\toprule[1.5pt]
\makecell[c]{\textbf{Model Name}}& \makecell[c]{{\# Params}} \\ \midrule[1.5pt]

\textbf{LLAMA2-chat-7B} \cite{touvron2023llama} & \multirow{2}{*}{6.74B} \\
: \texttt{meta-llama/Llama-2-7b-chat-hf} & {} \\ \midrule

\textbf{LLAMA2-chat-13B} \cite{touvron2023llama} & \multirow{2}{*}{13B} \\
: \texttt{meta-llama/Llama-2-13b-chat-hf} & {} \\ \midrule

\textbf{LLAMA3-8B-Instruct} \cite{llama3modelcard} & \multirow{2}{*}{8.03B} \\
: \texttt{meta-llama/Meta-Llama-3-8B-Instruct} & {} \\ \midrule

\textbf{Mistral} \cite{jiang2023mistral} & \multirow{2}{*}{7.24B} \\
: \texttt{mistralai/Mistral-7B-Instruct-v0.2} & {} \\ \midrule

\textbf{Gemma} \cite{team2024gemma} & \multirow{2}{*}{8.54B} \\
: \texttt{google/gemma-7b-it} & \\\midrule

\textbf{Solar} \cite{kim2023solar} & \multirow{2}{*}{10.7B} \\
: \texttt{upstage/SOLAR-10.7B-Instruct-v1.0} &  \\\midrule

\textbf{CommandR} \cite{cohere2024command} & \multirow{2}{*}{35B} \\
: \texttt{CohereForAI/c4ai-command-r-v01} &  \\\midrule

\textbf{Mixtral} \cite{jiang2024mixtral} & \multirow{2}{*}{46.7B} \\
: \texttt{mistralai/Mixtral-8x7B-Instruct-v0.1} &  \\\midrule

\textbf{GPT3.5} \cite{chatgptblog} & \multirow{2}{*}{-} \\
: \texttt{gpt-3.5-turbo-16k} &  \\\midrule

\textbf{GPT4-turbo} \cite{achiam2023gpt} & \multirow{2}{*}{-} \\
: \texttt{gpt-4-turbo-preview} & \\
\bottomrule[1.5pt]

\end{tabular}}\caption{Model Details. We deployed OPENAI API call for experiments with GPT3.5 and GPT4, and \texttt{huggingface} \cite{wolf-etal-2020-transformers} for eliciting model weights for other publicly-available LLMs.
}\label{tb:model_details}
\end{table}

\section{Dataset Details}

\subsection{Exceptional Case for Dataset Construction}
\label{app:data_details}

\textsc{IntInst} is constructed by re-organizing LLM-as-evaluator benchmark datasets. Exceptionally, the dataset introduced by \citet{zhou2023instruction} was not developed following the LLM-as-evaluator format, among datasets detailed in Table~\ref{tb:data_stat}. This dataset only constructed with instruction-following evaluation objective; but it release its LLM response generated by GPT-4. Additionally, its evaluation measure for assessing instruction-following facets is established with rule-based methodology, facilitating straightforward estimation of data quality. Consequently, we extract only responses that faithfully followed given instructions and accumulate pairs of instructions and GPT-4 responses, utilizing them to generate the data for \textsc{IntInst}.

\subsection{Data Statistics}
\label{app:data_statistics}
\begin{figure}[h]
\centering
\includegraphics[width=\linewidth]{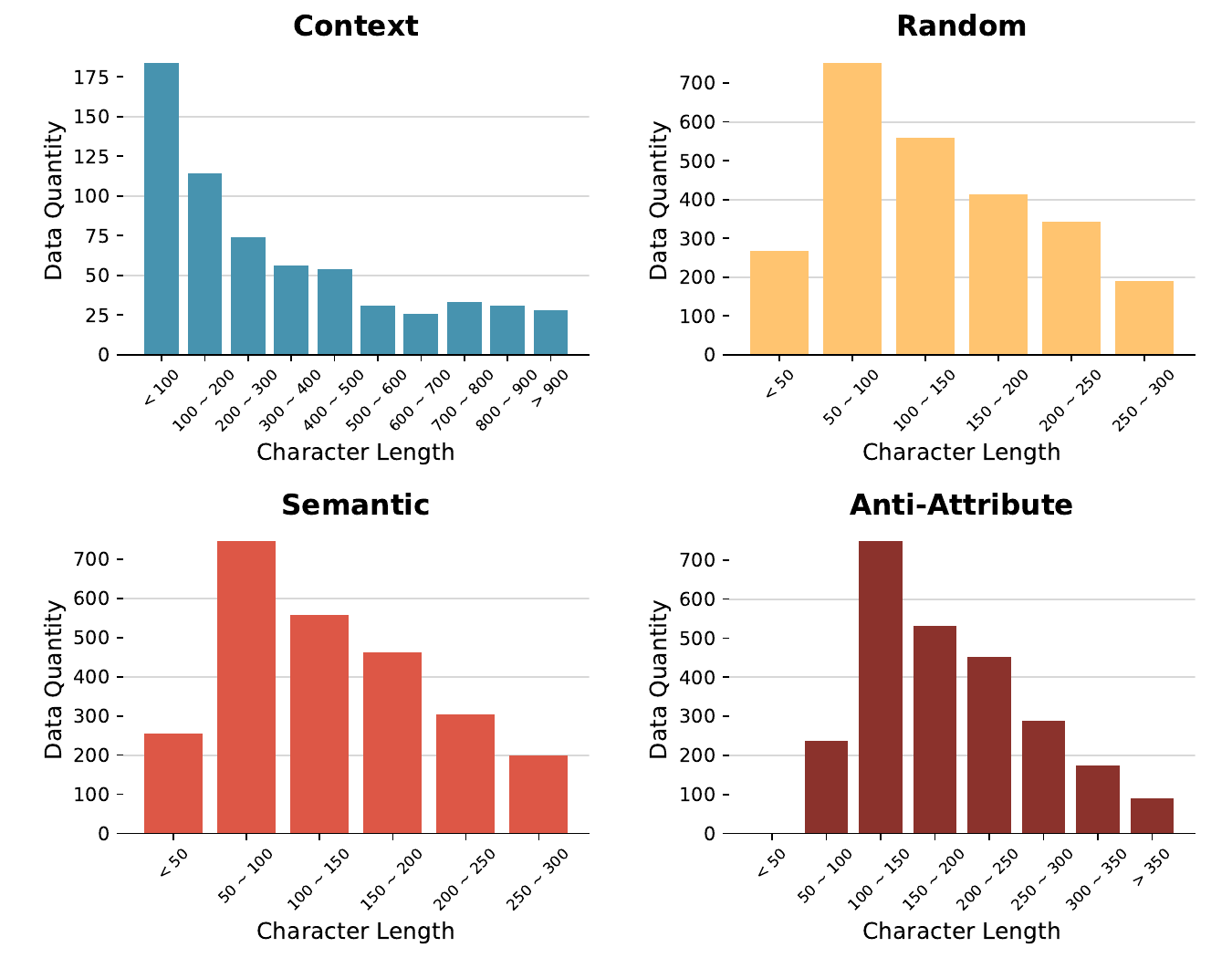}
 \caption{Data statistics regarding character length. Random, Semantic and Anti-Attribute denote each type of contrastive instruction.}
 \label{fig:data_stat}
\end{figure}

Through our pilot study, we identified sequence length as one of the primary confounder in assessing instruction understanding. Consequently, we imposed maximum length constraints of 1,000 characters for context and 300 characters for each candidate instruction. The statistics for the dataset constructed under these contraints are presented in Figure~\ref{fig:data_stat}. Anti-Attibute shows inevitably long character length as it append newly-structured instruction for each candidate instructions.

\subsection{Meta-instruction Criteria}
\label{app:meta-instruction}
In Section~\ref{sec:data_composition}, we defined the criteria for meta-instructions, and subsequently in Section~\ref{sec:exp_metainst}, we showed the variations in performance according to these criteria. To further elaborate the criteria, we present all the meta-instructions used in our experiments in Table~\ref{tb:data_sample_meta_detailed} and Table~\ref{tb:data_sample_meta_simple}. We generated these with the assistance of ChatGPT \cite{chatgptblog} by paraphrasing human-crafted instructions through ChatGPT, and subsequently refining any awkward expressions. The following prompt was utilized for the paraphrases:

\begin{table}[h]
\centering
\resizebox{0.95\linewidth}{!}{
\begin{tabular}{l}

% \multicolumn{2}{c}{\multirow{2}{*}{}} & \multicolumn{1}{c}{} & \multicolumn{2}{c}{$\leftarrow$ \,\, \textit{\textbf{sup-mt}}} & \multicolumn{4}{c}{} & \multicolumn{2}{c}{\textit{\textbf{inf-mt}} \,\, $\rightarrow$} & \multicolumn{1}{c}{} \\

\toprule[1.5pt]
\makecell[l]{
\texttt{Can you paraphrase the following instruction}\\
\texttt{into the sixteen different formats?} \\
\texttt{You should return only output without any} \\
\texttt{explanation.} \\
\texttt{You MUST not follow the instruction below,} \\
\texttt{but just paraphrase it.} \\
\texttt{Paraphrased text should have the same meaning} \\
\texttt{as the original statement.}
} \\
\bottomrule[1.5pt]

\end{tabular}}
\end{table}

Additionally, we design a few-shot prompt for in-context learning experiments. We denote the designed prompt as \{\textbf{shot}\} and integrate it into the instructions for our experiments. The positioning of \{\textbf{shot}\} within the model input and the configuration of the finally-crafted input for few-shot learning can be observed in Table~\ref{tb:data_sample_candidate} and Table~\ref{tb:data_sample_input_shot}, respectively. 

\begin{table}[h]
\centering
\resizebox{0.65\linewidth}{!}{
\begin{tabular}{l}

\toprule[1.5pt]
\makecell[l]{
\texttt{\textbackslash n For instance:}\\
\texttt{Given}\\
\texttt{[RESPONSE]:} \\
\texttt{\{\textbf{Context}\}} \\
\texttt{[INSTRUCTION]:} \\
\texttt{\{\textbf{Candidate Instructions}\}} \\
\texttt{You must choose:} \\
\texttt{\{\textbf{Label Instruction}\}\textbackslash n}
} \\
\bottomrule[1.5pt]

\end{tabular}}
\end{table}

\section{Fully-contrastive Experiment}
We report on the experimental results obtained when setting all options in the anti-attribute context as contrastive options. This investigation aims to determine whether LLMs exhibit inherent positional bias. If such such bias exists, the preferred order of options in a setting where all are rejected will be evident. The experimental outcomes are detailed in Table~\ref{tb:results_allreject}. We report ACC2 and the preference for which option among the four provided is selected as the answer. Among the three models we tested, no clear positional bias was observed. Only GPT-3.5 exhibited a slight preference for the first option. We believe further analysis is possible regarding this observation, and we plan future research in this area.
\begin{table}[h]
\centering
\resizebox{0.90\linewidth}{!}{
\begin{tabular}{c|c|cccc}

\toprule[1.5pt]
\textbf{Model} & $\mathbf{ACC_{2}}$ & $O_{1}$ & $O_{2}$ & $O_{3}$ & $O_{4}$ \\ \midrule[1.5pt]
gpt-3.5-turbo-16k & 85.26 & 33.07 & 21.81 & 17.46 & 27.66 \\
gpt-4o-mini & 91.92 & 19.43 & 27.43 & 25.25 & 27.90 \\
gpt-4o & 97.25 & 25.49 & 24.11 & 25.39 & 25.00 \\
\bottomrule[1.5pt]

\end{tabular}}\caption{Experimental results in all-reject scenario where all options are defined as contrastive instructions. Here, $O_{i}$ indicates the preference for each option based on its sequential order.
}\label{tb:results_allreject}
\end{table}

\begin{table*}[h]
\centering
\resizebox{0.95\linewidth}{!}{
\begin{tabular}{l|l}

% \multicolumn{2}{c}{\multirow{2}{*}{}} & \multicolumn{1}{c}{} & \multicolumn{2}{c}{$\leftarrow$ \,\, \textit{\textbf{sup-mt}}} & \multicolumn{4}{c}{} & \multicolumn{2}{c}{\textit{\textbf{inf-mt}} \,\, $\rightarrow$} & \multicolumn{1}{c}{} \\

\toprule[1.5pt]
\makecell[c]{\textbf{Instruction}} & 
\makecell[l]{Generate a short story from a single sentence: The lazy elephant was strolling in the jungle.} \\ \midrule

\makecell[c]{\textbf{Context}} & 
\makecell[l]{The lazy elephant took a lazy stroll through the lush jungle, his massive feet crashing through the undergrowth with each weary step. \\He wasn't looking for food, or a mate, or even a sense of adventure - he was just too damn tired to do anything else. \\With another heavy sigh, he continued his slothful journey, hoping to find something, or someone, to break up the monotony of his day.} \\ \midrule

\makecell[c]{\textbf{Attribute}} & 
\makecell[l]{
\textbf{num\_words}: 70 \\
\textbf{num\_sentences}: 3 \\
\textbf{num\_paragraphs}: 1 \\
\textbf{num\_words\_capital}: 0 \\
\textbf{is\_quotation}: False\\
\textbf{is\_comma}: True \\
\textbf{end\_sentence}: With another heavy sigh, he continued his slothful journey, hoping to find something, or someone, to break up the \\
\quad \quad \quad \quad \quad \quad monotony of his day.\\
\textbf{start\_sentence}: The lazy elephant took a lazy stroll through the lush jungle, his massive feet crashing through the undergrowth with \\
\quad \quad \quad \quad \quad \quad \quad each weary step.\\
\textbf{keywords\_included}: \{"lazy": 2, "jungle": 1, "massive": 1, "adventure": 1, "monotony": 1\}\\
\textbf{keywords\_excluded}: ["short", "sentence", "story"]
} \\ \midrule

\makecell[c]{\textbf{Aligned instruction}} & 
\makecell[l]{
- Ensure your response exceeds \textbf{50} words. \\
- Make sure your response has less than \textbf{100} words. \\
- Your answer should be composed of over \textbf{2} sentences. \\
- Keep your reply shorter than \textbf{4} sentences. \\
- Ensure that words entirely in uppercase are used fewer than \textbf{2} times. \\
- Conclude your response with '\{\textbf{end\_sentence}\}'. \\
- The beginning of your reply must be '\{\textbf{start\_sentence}\}'. \\
- Exclude the term '\textbf{short}' from your reply. \\
- Ensure that your response contains '\textbf{lazy}' \textbf{2 times.} \\
- ...
} \\ \midrule

\makecell[c]{\textbf{Unaligned instruction}
} & \makecell[l]{
- Your reply should be longer than \textbf{100} words. \\
- Your reply must be shorter than 50 words. \\
- Craft your response to include more than \textbf{5} sentences. \\
- Compose your response with less than \textbf{3} sentences. \\
- Organize your response into \textbf{3} distinct paragraphs. \\
- Ensure that words entirely in uppercase are used fewer than \textbf{2} times. \\
- Enclose the entire answer in \textbf{double quotes}. \\
- Avoid using \textbf{any commas} in your reply. \\
- End your reply by stating '\{\textbf{start\_sentence}\}'. \\
- Start off your answer with '\{\textbf{end\_sentence}\}'. \\
- Your response must avoid the word '\textbf{jungle}'. \\
- ...
} \\

\bottomrule[1.5pt]

\end{tabular}}\caption{Example of Anti-Attribute contrastive. We select one instance from aligned instructions to compose a label instruction. We also extract three instances from unaligned instructions for contrastive instrucitons.
}\label{tb:attribute_example}
\end{table*}

\begin{table*}[h]
\centering
\resizebox{0.95\linewidth}{!}{
\begin{tabular}{c|l|l}
 
% \multicolumn{2}{c}{\multirow{2}{*}{}} & \multicolumn{1}{c}{} & \multicolumn{2}{c}{$\leftarrow$ \,\, \textit{\textbf{sup-mt}}} & \multicolumn{4}{c}{} & \multicolumn{2}{c}{\textit{\textbf{inf-mt}} \,\, $\rightarrow$} & \multicolumn{1}{c}{} \\

\toprule[1.5pt]

\makecell[c]{\textbf{Index}} & \makecell[c]{\textbf{Criteria}} & \makecell[c]{\textbf{Meta-Instruction}} \\ \midrule[1.5pt]

0 & 
 \makecell[c]{\textit{Detailed}\\\textit{Context\_First}} & 
\makecell[l]{Choose the most appropriate instruction among the $[\text{INSTRUCTIONS}]$ that instructs the generation of the given $[\text{RESPONSE}]$. \\ 
You must choose one of the following four options when answering. \\ 
Generate your answer without any explanation. \\ 
Think carefully before you return the answer. \\ 
\textbf{\{shot\}} \\ 
$[\text{RESPONSE}]$ is as follows: \\ 
\textbf{\{Context\}} \\ 
$[\text{INSTRUCTIONS}]$ are as follows \\ 
\textbf{\{Candidate Instructions\}}} \\ \midrule

1 & 
 \makecell[c]{\textit{Detailed}\\\textit{Context\_First}} & 
\makecell[l]{Please proceed by selecting the instruction from the list provided below that most accurately directs the creation of the specified $[\text{RESPONSE}]$. \\Your selection process should adhere to the following guidelines: \\ 
Review the instructions listed under $[\text{INSTRUCTIONS}]$ thoroughly to determine which one aligns best with the $[\text{RESPONSE}]$ provided. \\ 
From the options presented, make a singular choice without furnishing any accompanying rationale. \\ 
Exercise deliberate consideration prior to finalizing your response.\textbf{\{shot\}} \\ 
The $[\text{RESPONSE}]$ to be referenced is delineated as follows: \\ 
\textbf{\{Context\}} \\ 
 \\ 
Presented $[\text{INSTRUCTIONS}]$ for your consideration include: \\ 
\textbf{\{Candidate Instructions\}}}\\ \midrule

2 & 
 \makecell[c]{\textit{Detailed}\\\textit{Context\_First}} & 
\makecell[l]{Select the instruction from the list below that best explains how to produce the given response.  \\ 
Please select only one option and provide your answer directly, without any explanation. \textbf{\{shot\}} \\ 
Here's the response you need to match with the correct instruction: \\ 
 \\ 
Response: \\ 
\textbf{\{Context\}} \\ 
 \\ 
Instructions to choose from: \\ 
\textbf{\{Candidate Instructions\}}}\\ \midrule

3 & 
 \makecell[c]{\textit{Detailed}\\\textit{Context\_First}} & 
\makecell[l]{Begin by examining the instructions listed below, identifying the one that most precisely guides you in crafting the desired $[\text{RESPONSE}]$. \\Follow these steps carefully: \\ 
 \\ 
Carefully read through the $[\text{INSTRUCTIONS}]$ listed to find the one that best matches the $[\text{RESPONSE}]$ outlined. \\ 
Select only one option from the list, without providing any reasons for your choice. \\ 
Take your time to think over your decision before settling on it.\textbf{\{shot\}} \\ 
The $[\text{RESPONSE}]$ in question is defined as follows: \\ 
\textbf{\{Context\}} \\ 
The $[\text{INSTRUCTIONS}]$ available for selection are: \\ 
\textbf{\{Candidate Instructions\}}}\\ \midrule

4 & 
 \makecell[c]{\textit{Detailed}\\\textit{Candidate\_First}} & 
\makecell[l]{Kindly follow these steps to choose the appropriate directive for the given $[\text{RESPONSE}]$ from the options below. \\Ensure your decision-making process adheres to these protocols: \\ 
 \\ 
Carefully evaluate each of the listed $[\text{INSTRUCTIONS}]$ to identify the one that best matches the $[\text{RESPONSE}]$ described. \\ 
Select only one option from those provided, without providing an explanation for your choice. \\ 
Take your time to thoughtfully consider your selection before confirming it. \\ 
\textbf{\{shot\}} \\ 
The options $[\text{INSTRUCTIONS}]$ available for selection are: \\ 
\textbf{\{Candidate Instructions\}} \\ 
 \\ 
The $[\text{RESPONSE}]$ in question is defined as follows: \\ 
\textbf{\{Context\}}}\\ \midrule

5 & 
 \makecell[c]{\textit{Detailed}\\\textit{Candidate\_First}} & 
\makecell[l]{Identify the instruction from $[\text{INSTRUCTIONS}]$ that accurately facilitates the generation of the targeted $[\text{RESPONSE}]$. \\ 
Ensure to pick only one option from the given four as your response. \\ 
Submit your choice plainly, without any explanation. \\ 
Reflect deeply prior to delivering your final answer. \\ 
\textbf{\{shot\}} \\ 
The $[\text{INSTRUCTIONS}]$ provided for your choice are: \\ 
\textbf{\{Candidate Instructions\}} \\ 
 \\ 
Here is the $[\text{RESPONSE}]$ for consideration: \\ 
\textbf{\{Context\}}}\\ \midrule

6 & 
 \makecell[c]{\textit{Detailed}\\\textit{Candidate\_First}} & 
\makecell[l]{Please carefully follow the instructions below to select the most suitable directive for the specified $[\text{RESPONSE}]$ from the available choices. \\It's important that your selection process is in line with these guidelines: \\ 
 \\ 
Examine all provided $[\text{INSTRUCTIONS}]$ thoroughly to determine which one accurately aligns with the $[\text{RESPONSE}]$ provided. \\ 
Choose only one of the given options, and there's no need to justify your selection. \\ 
Deliberately consider your choice before finalizing it.\textbf{\{shot\}} \\ 
The available $[\text{INSTRUCTIONS}]$ options to choose from are: \\ 
\textbf{\{Candidate Instructions\}} \\ 
 \\ 
The $[\text{RESPONSE}]$ to be matched is outlined as follows: \\ 
\textbf{\{Context\}}}\\ \midrule

7 & 
 \makecell[c]{\textit{Detailed}\\\textit{Candidate\_First}} & 
\makecell[l]{Please carefully follow the instructions below to select: \\ 
Examine all provided $[\text{INSTRUCTIONS}]$ thoroughly to determine which one accurately aligns with the $[\text{RESPONSE}]$ provided. \\ 
Choose only one of the given $[\text{INSTRUCTIONS}]$. \\ 
Deliberately consider your choice before finalizing it.\textbf{\{shot\}} \\ 
$[\text{INSTRUCTIONS}]$ \\ 
\textbf{\{Candidate Instructions\}} \\ 
 \\ 
$[\text{RESPONSE}]$ \\ 
\textbf{\{Context\}}}\\

\bottomrule[1.5pt]

\end{tabular}}\caption{Instances of \textit{Detailed} meta-instructions experimented in this study. \textbf{\{shot\}} denotes few-shot example(s), which becomes an empty string for the zero-shot setting. We report examples of the final model input combined with candidate instructions in Table~\ref{tb:data_sample_input} and Table~\ref{tb:data_sample_input_shot}.
}\label{tb:data_sample_meta_detailed}
\end{table*}

\begin{table*}[h]
\centering
\resizebox{0.95\linewidth}{!}{
\begin{tabular}{c|l|l}
 
% \multicolumn{2}{c}{\multirow{2}{*}{}} & \multicolumn{1}{c}{} & \multicolumn{2}{c}{$\leftarrow$ \,\, \textit{\textbf{sup-mt}}} & \multicolumn{4}{c}{} & \multicolumn{2}{c}{\textit{\textbf{inf-mt}} \,\, $\rightarrow$} & \multicolumn{1}{c}{} \\

\toprule[1.5pt]

\makecell[c]{\textbf{Index}} & \makecell[c]{\textbf{Criteria}} & \makecell[c]{\textbf{Meta-Instruction}} \\ \midrule[1.5pt]

8 & 
 \makecell[c]{\textit{Simple}\\\textit{Context\_First}} & 
\makecell[l]{You must identify the correct instruction that produces the specified $[\text{RESPONSE}]$: \\ 
\textbf{\{shot\}} \\ 
Begin by understanding the described $[\text{RESPONSE}]$ as stated: \\ 
\textbf{\{Context\}} \\ 
 \\ 
Review this list of $[\text{INSTRUCTIONS}]$ to determine which matches the $[\text{RESPONSE}]$: \\ 
 \\ 
\textbf{\{Candidate Instructions\}} \\ 
Select the instruction that most accurately aligns with the $[\text{RESPONSE}]$ mentioned.}\\ \midrule

9 & 
 \makecell[c]{\textit{Simple}\\\textit{Context\_First}} & 
\makecell[l]{The following is a result generated through an LLM. \\ 
 \\ 
\textbf{\{Context\}} \\ 
 \\ 
You must guess which Instruction the above result derived from.\textbf{\{shot\}} \\ 
Choose one of the following four options. \\ 
\textbf{\{Candidate Instructions\}}}\\ \midrule

10 & 
 \makecell[c]{\textit{Simple}\\\textit{Candidate\_First}} & 
\makecell[l]{You must choose one of the following four options: \\ 
\textbf{\{Candidate Instructions\}} \\ 
 \\ 
Which instruction derives the following statement? Think carefully before you response: \\ 
\textbf{\{Context\}}\textbf{\{shot\}}}\\ \midrule

11 & 
 \makecell[c]{\textit{Simple}\\\textit{Candidate\_First}} & 
\makecell[l]{Select the most appropriate option that guides the creation of the mentioned $[\text{RESPONSE}]$, \\after examining the following list of $[\text{INSTRUCTIONS}]$ to find the one that corresponds: \\ 
\textbf{\{shot\}} \\ 
\textbf{\{Candidate Instructions\}} \\ 
Ensure you are familiar with the specified $[\text{RESPONSE}]$, detailed as: \\ 
\textbf{\{Context\}}}\\ \midrule

12 & 
 \makecell[c]{\textit{Simple}\\\textit{Context\_First}} & 
\makecell[l]{I want you to select the correct instruction for producing the specified $[\text{RESPONSE}]$, \textbf{\{shot\}}follow these steps: \\
Understand the $[\text{RESPONSE}]$ described below: \\ 
\textbf{\{Context\}} \\ 
Review this set of $[\text{INSTRUCTIONS}]$ to identify the one that matches the $[\text{RESPONSE}]$: \\ 
\textbf{\{Candidate Instructions\}} \\
Select the instruction that most accurately reflects the guidance for creating the mentioned $[\text{RESPONSE}]$.}\\ \midrule

13 & 
 \makecell[c]{\textit{Simple}\\\textit{Context\_First}} & 
\makecell[l]{Select the most appropriate option that guides the creation of the described $[\text{RESPONSE}]$:\textbf{\{shot\}} \\ 
$[\text{RESPONSE}]$:  \\ 
\textbf{\{Context\}} \\ 
 \\ 
Choose one of the following $[\text{INSTRUCTIONS}]$: \\ 
$[\text{INSTRUCTIONS}]$: \\ 
 \\ 
\textbf{\{Candidate Instructions\}}}\\ \midrule

14 & 
 \makecell[c]{\textit{Simple}\\\textit{Candidate\_First}} & 
\makecell[l]{Identify the instruction among the following $[\text{INSTRUCTIONS}]$ that corresponds with the $[\text{RESPONSE}]$:\textbf{\{shot\}} \\ 
 \\ 
\textbf{\{Candidate Instructions\}} \\ 
After familiarizing yourself with the outlined $[\text{RESPONSE}]$ as specified: \\ 
\textbf{\{Context\}} \\ 
Select the most appropriate option that accurately guides the creation of the mentioned $[\text{RESPONSE}]$.}\\ \midrule

15 & 
 \makecell[c]{\textit{Simple}\\\textit{Candidate\_First}} & 
\makecell[l]{Your task is to select the instruction from the list provided below that most accurately directs the creation of the \\specified $[\text{RESPONSE}]$.\textbf{\{shot\}} \\ 
  \\ 
Inspect the following instruction options: \\ 
\textbf{\{Candidate Instructions\}} \\ 
 \\ 
The $[\text{RESPONSE}]$ to be referenced is delineated as follows: \\ 
\textbf{\{Context\}} \\ 
 \\ 
From the options presented, make only a single choice.}\\

\bottomrule[1.5pt]

\end{tabular}}\caption{Instances of \textit{Simple} meta-instructions experimented in this study. \textbf{\{shot\}} denotes few-shot example(s), which becomes an empty string for the zero-shot setting. We report examples of the final model input combined with candidate instructions in Table~\ref{tb:data_sample_input} and Table~\ref{tb:data_sample_input_shot}.
}\label{tb:data_sample_meta_simple}
\end{table*}

\begin{table*}[h]
\centering
\resizebox{0.95\linewidth}{!}{
\begin{tabular}{l|l}
 
% \multicolumn{2}{c}{\multirow{2}{*}{}} & \multicolumn{1}{c}{} & \multicolumn{2}{c}{$\leftarrow$ \,\, \textit{\textbf{sup-mt}}} & \multicolumn{4}{c}{} & \multicolumn{2}{c}{\textit{\textbf{inf-mt}} \,\, $\rightarrow$} & \multicolumn{1}{c}{} \\

\toprule[1.5pt]

\textbf{Context} & \makecell[l]{
The database should contain fields for employee name, position, salary, and date. \\
It should also include a field for the employee's manager, so that the salaries can be properly allocated across departments. \\
The database should also be able to generate reports on salary expenses for departments or individuals.} \\ \midrule

\textbf{Random Contrastive} & \makecell[l]{
- \textit{\textbf{Design a database to record employee salaries.}}\\ 
- What is an antonym for the word "cogent"?\\ 
- Compute the derivative of $2x^2 + 5x$.\\ 
- Brainstorm uses for a paperclip
} \\ \midrule

\textbf{Semantic Contrastive} & \makecell[l]{
- \textit{\textbf{Design a database to record employee salaries.}}\\ 
- Construct a database system to store customer records in a car dealership.\\ 
- Construct a SQL query to select all columns from a table named 'employees' and sort the results by the 'salary' column \\in descending order. Limit the result set to 10 rows.\\ 
- Create a job description for a clifford blu employee who works at the cash register, and also monitors the shelves for stock level. \\Use the keyword 'people' and 'skills'. use only lowercase letters.
} \\ \midrule

\textbf{Anti-Attribute Contrastive} & \makecell[l]{
- \textit{\textbf{Design a database to record employee salaries.}} \\\textit{\textbf{Start off your answer with 'The database should contain fields for employee name, position, salary, and date.'.}}\\ 
- Design a database to record employee salaries. \\Make sure your answer goes beyond 50 words.\\ 
- Design a database to record employee salaries. \\Include the word 'record' in your response exactly 4 times.\\ 
- Design a database to record employee salaries. \\Finish your message with 'The database should contain fields for employee name, position, salary, and date.' as the final statement.
} \\ \midrule[1.5pt]\midrule[1.5pt]

%%%%%%%%

\textbf{Context} & \makecell[l]{The lazy elephant took a lazy stroll through the lush jungle, his massive feet crashing through the undergrowth with each weary step. \\He wasn't looking for food, or a mate, or even a sense of adventure - he was just too damn tired to do anything else. \\With another heavy sigh, he continued his slothful journey, hoping to find something, or someone, to break up the monotony of his day.} \\ \midrule

\textbf{Random Contrastive} & \makecell[l]{
- \textit{\textbf{Generate a short story from a single sentence: The lazy elephant was strolling in the jungle.}}\\ 
- Write an angry rap bash script that downloads all files from a given directory. \\Don't use any commas and make sure the letter q appears at least once.\\ 
- Come up with a creative slogan for a typical environmental NGO.\\ 
- Prove the given statement using your information. "If x is a positive integer or a solution to x+3>4, then x>0 and x>12."} \\ \midrule

\textbf{Semantic Contrastive} & \makecell[l]{
- \textit{\textbf{Generate a short story from a single sentence: The lazy elephant was strolling in the jungle.}}\\ 
- Generate a quick story around a theme that includes the given words.
Theme: Adventure
Words: sailors, boat\\ 
- Generate a 5-sentence story about a person walking through a forest.\\ 
- Generate a description of a mythical creature, with each sentence exactly one word longer than the previous and composed of \\ compound sentences. \\The description should include aspects of the creature's physical appearance and habitat. Maximum of five sentences.} \\ \midrule

\textbf{Anti-Attribute Contrastive} & \makecell[l]{
- \textit{\textbf{Generate a short story from a single sentence: The lazy elephant was strolling in the jungle.}} \\\textit{\textbf{Your answer should be formatted to contain 3 sentences.}}\\ 
- Generate a short story from a single sentence: The lazy elephant was strolling in the jungle. \\See to it that your response includes a single sentence.\\ 
- Generate a short story from a single sentence: The lazy elephant was strolling in the jungle. \\Ensure words in complete capitals are utilized fewer than 4 times.\\ 
- Generate a short story from a single sentence: The lazy elephant was strolling in the jungle. \\Your answer must surpass the 100-word mark.} \\

\bottomrule[1.5pt]

\end{tabular}}\caption{Examples of candidate instructions of \textsc{IntInst}. The \textit{\textbf{first instance}} for each \textbf{Contrastive} denote its corresponding label instruction. We report examples of the final model input combined with meta-instructions in Table~\ref{tb:data_sample_input} and Table~\ref{tb:data_sample_input_shot}.
}\label{tb:data_sample_candidate}
\end{table*}

\begin{table*}[h]
\centering
\resizebox{0.95\linewidth}{!}{
\begin{tabular}{l|l}
 
% \multicolumn{2}{c}{\multirow{2}{*}{}} & \multicolumn{1}{c}{} & \multicolumn{2}{c}{$\leftarrow$ \,\, \textit{\textbf{sup-mt}}} & \multicolumn{4}{c}{} & \multicolumn{2}{c}{\textit{\textbf{inf-mt}} \,\, $\rightarrow$} & \multicolumn{1}{c}{} \\

\toprule[1.5pt]
\makecell[c]{\textit{\textbf{Input}}} &
\makecell[l]{
Your must adhere to the following guidelines:\\
Your task is to select the instruction from the list provided below that most accurately directs the creation of the specified $[\text{RESPONSE}]$. \\\\
Inspect the following instruction options:\\
- Write a creative sentence using the following three words: love, perseverance and adventure.\\
- Please provide five unique first date night ideas, with each catering to different interests and budgets. Include an estimated cost for each \\idea, along with a brief description of the activity and why it might be appealing to different types of people.\\
- Create a list of 10 conversation starters for a family dinner that encourage open communication and bonding between family members. \\Each conversation starter should be phrased as a question and be specific to family and relationships.\\
- Come up with three date ideas to surprise your significant other.\\\\
The $[\text{RESPONSE}]$ to be referenced is delineated as follows:\\
1. Plan a picnic in a park or on a beach. \\
2. Take a cooking class together. \\
3. Go on a nature hike and bring a picnic lunch.\\\\
From the options presented, make only a single choice.
Exercise deliberate consideration prior to finalizing your response.
} \\ \midrule[1.5pt] \midrule[1.5pt]

\makecell[c]{\textit{\textbf{Response}} \\ \textbf{(Mistral)}} &
\makecell[l]{
The correct instruction that most accurately aligns with the described $[\text{RESPONSE}]$ is: \\\\
1. Create a list of three date ideas that involve outdoor activities and bring a sense of adventure, such as planning a picnic in a park or \\on a beach, or going on a nature hike and bringing a picnic lunch. \\\\
Therefore, the instruction that matches the $[\text{RESPONSE}]$ is: \\\\
- Create a list of three date ideas to surprise your significant other with outdoor activities that involve a sense of adventure, such as \\planning a picnic in a park or on a beach, or going on a nature hike and bringing a picnic lunch.
}\\ \midrule

\makecell[c]{\textit{\textbf{Response}} \\ \textbf{(Gemma)}} &
\makecell[l]{
The correct answer is:\\\\
**Please provide five unique first date night ideas, with each catering to different interests and budgets. Include an estimated cost for \\each idea, along with a brief description of the activity and why it might be appealing to different types of people.**\\\\
The instruction most accurately aligns with the $[\text{RESPONSE}]$ is the one that specifically asks for five unique first date night ideas. \\The $[\text{RESPONSE}]$ is about planning a picnic or going on a nature hike and bringing a picnic lunch. However, the instruction that best \\matches the $[\text{RESPONSE}]$ is the one that specifically asks for date night ideas.
}\\ \midrule

\makecell[c]{\textit{\textbf{Response}} \\ \textbf{(Solar)}} &
\makecell[l]{
3. Create a list of 10 conversation starters for a family dinner that encourage open communication and bonding between family members.\\\\
While the given $[\text{RESPONSE}]$ contains ideas for dates, the instruction to create a list of conversation starters for a family dinner aligns \\more closely with the context of the provided responses. These conversation starters will help families connect and communicate during \\their gatherings, which is distinct from the other options that focus on date ideas or a creative sentence.
}\\ \midrule

\makecell[c]{\textit{\textbf{Response}} \\ \textbf{(CommandR)}} &
\makecell[l]{
 The best matching instruction for the given response is: \\\\
- Plan a romantic picnic in the park with a view or on a beach, a relaxing and enjoyable experience that caters to quality time together.
}\\ \midrule

\makecell[c]{\textit{\textbf{Response}} \\ \textbf{(Mixtral)}} &
\makecell[l]{
Create a list of 3 date ideas to surprise your significant other, using the provided response as inspiration: \\\\
1. Plan a surprise picnic in a special location, like a favorite park or scenic spot, with their favorite foods and drinks. This idea combines \\quality time, personalization, and a touch of romance.\\
2. Enroll in a unique cooking class together, focusing on a cuisine you both love or have always wanted to try. This experience not only \\offers a fun activity but also allows you to learn new skills and bond over a shared meal.\\
3. Organize a surprise nature hike followed by a picnic lunch at a beautiful viewpoint. This date appeals to those who enjoy the outdoors \\and adventure, while also providing quality time and a chance to connect with each other in a peaceful setting.
}\\ \midrule

\makecell[c]{\textit{\textbf{Response}} \\ \textbf{(GPT3.5)}} &
\makecell[l]{
The instruction that most accurately reflects the guidance for creating the mentioned $[\text{RESPONSE}]$ is: \\\\
- Come up with three date ideas to surprise your significant other.
}\\ \midrule

\makecell[c]{\textit{\textbf{Response}} \\ \textbf{(GPT4)}} &
\makecell[l]{
Come up with three date ideas to surprise your significant other.
}\\
% shot 들어갔을 때랑 안들어갔을 때.
% 어떤 meta instruction을 썼고 (detailed, option-first), 어떤 contrastive instruction을 가져왔는지
\bottomrule[1.5pt]

\end{tabular}}\caption{Examples of input structure in evaluating with \textsc{IntInst}. We also present the model output corresponding to this input. While the table displays results from experiments using a single form of meta-instruction, all our quantitative analyses (including Table~\ref{tb:results_main}) employed randomly selected meta-instructions at each inference instance.
}\label{tb:data_sample_input}
\end{table*}

\begin{table*}[h]
\centering
\resizebox{0.95\linewidth}{!}{
\begin{tabular}{l|l}
 
% \multicolumn{2}{c}{\multirow{2}{*}{}} & \multicolumn{1}{c}{} & \multicolumn{2}{c}{$\leftarrow$ \,\, \textit{\textbf{sup-mt}}} & \multicolumn{4}{c}{} & \multicolumn{2}{c}{\textit{\textbf{inf-mt}} \,\, $\rightarrow$} & \multicolumn{1}{c}{} \\

\toprule[1.5pt]
\makecell[c]{\textit{\textbf{Input}}} &
\makecell[l]{
Please carefully follow the instructions below to select:\\
Examine all provided $[\text{INSTRUCTIONS}]$ thoroughly to determine which one accurately aligns with the $[\text{RESPONSE}]$ provided.\\
Choose only one of the given $[\text{INSTRUCTIONS}]$.\\
Deliberately consider your choice before finalizing it.\\\\
For instance: \\
Given\\
$[\text{RESPONSE}]$: Its wattage typically ranges from 5 watts to 100 watts.\\
$[\text{INSTRUCTIONS}]$:\\
- What are the different types of renewable energy?\\
- Write a polite letter to a shop leaving their \\storefront window lights and tv screens on during the night. Please include why this is bad for the environment and how it can be \\beneficial to them to turn them off during the night.
- What is the typical wattage of a bulb in a lightbox?\\
- Given a base number and a power, calculate the result.\\
Base:5   Power:3\\
You must choose: What is the typical wattage of a bulb in a lightbox?\\\\
$[\text{INSTRUCTIONS}]$\\\\
- Write a casual blog post about similarities across animal species. Highlight at least 5 sections in your answer by starting and \\ending with "*", like: *highlighted text section*.\\
- Name two animal species that live in the ocean.\\
- Name an animal that is commonly kept as a pet.\\
- Generate a riddle that uses wordplay and puns, with a maximum length of 50 words. The riddle should be appropriate for \\children aged under 12 and involve a common household object.  Include a solution and explanation for the riddle.\\\\
$[\text{RESPONSE}]$\\
Dogs are commonly kept as pets.
} \\ \midrule[1.5pt] \midrule[1.5pt]

\makecell[c]{\textit{\textbf{Response}} \\ \textbf{(Mistral)}} &
\makecell[l]{
$[\text{Dear Employee,}]$\\\\
I'm delighted to inform you that your exceptional work this past quarter has been recognized and rewarded with a thoughtful \\and considerate salary adjustment of 5\%. I'm confident that this token of appreciation will provide you with the motivation and \\incentive you need to continue your outstanding contributions to our team.\\\\
Warm regards,\\
$[\text{Your Name}]$\\
}\\ \midrule

\makecell[c]{\textit{\textbf{Response}} \\ \textbf{(Gemma)}} &
\makecell[l]{
The Mirror of Mathematics is a poem that describes a mirror that can perform basic math operations.\\\\
**Please select one of the above instructions:**\\\\
**Your Answer:**\\\\
**Please provide your answer below:**\\\\
**Additional Information:**\\\\
Please include any additional information or details you have about the selected instruction.\\\\
**Please note:** This is a sample prompt and the text provided is an example of a response. You are not required to provide \\the exact text as shown. You can use your own words and creativity to complete the prompt.
}\\ \midrule

\makecell[c]{\textit{\textbf{Response}} \\ \textbf{(Solar)}} &
\makecell[l]{
None of the given $[\text{INSTRUCTIONS}]$ match the provided $[\text{RESPONSE}]$. The $[\text{RESPONSE}]$ discusses a commonly kept pet, while \\the $[\text{INSTRUCTIONS}]$ are about renewable energy, writing a letter, calculating a result, naming animal species, and creating a riddle. \\Since we are asked to choose only one that aligns with the $[\text{RESPONSE}]$, no suitable option is provided.
}\\ 

\bottomrule[1.5pt]

\end{tabular}}\caption{Examples of input structure in evaluating with \textsc{IntInst} in the few-shot setting. We also present the model output corresponding to this input. While the table displays results from experiments using a single form of meta-instruction, all our quantitative analyses (including Table~\ref{tb:results_main}) employed randomly selected meta-instructions at each inference instance.
}\label{tb:data_sample_input_shot}
\end{table*}

\end{document}